\documentclass{article}

\usepackage{microtype}
\usepackage{graphicx}
\usepackage{subfigure}
\usepackage{booktabs} 

\usepackage{hyperref}

\usepackage[accepted]{icml2025}

\usepackage{amsmath}
\usepackage{amssymb}
\usepackage{mathtools}
\usepackage{amsthm}

\usepackage{tabularx}

\usepackage[capitalize,noabbrev]{cleveref}

\theoremstyle{plain}

\theoremstyle{definition}

\theoremstyle{remark}

\usepackage[textsize=tiny]{todonotes}

\icmltitlerunning{Explain Like I'm Five: Using LLMs to Improve PDE Surrogate Models with Text}

\begin{document}

\twocolumn[
\icmltitle{Explain Like I'm Five: Using LLMs to Improve PDE Surrogate Models with Text}



\icmlsetsymbol{equal}{*}

\begin{icmlauthorlist}
\icmlauthor{Cooper Lorsung}{meche}
\icmlauthor{Amir Barati Farimani}{meche,mld,cheme,bme}
\end{icmlauthorlist}

\icmlaffiliation{meche}{Department of Mechanical Engineering, Carnegie Mellon University, Pittsburgh Pennsylvania, USA}
\icmlaffiliation{mld}{Machine Learning Department, Carnegie Mellon University, Pittsburgh Pennsylvania, USA}
\icmlaffiliation{cheme}{Machine Learning Department, Carnegie Mellon University, Pittsburgh Pennsylvania, USA}
\icmlaffiliation{bme}{Machine Learning Department, Carnegie Mellon University, Pittsburgh Pennsylvania, USA}

\icmlcorrespondingauthor{Amir Barati Farimani}{barati@cmu.edu}

\icmlkeywords{Machine Learning, ICML}

\vskip 0.3in
]



\printAffiliationsAndNotice{}  

\begin{abstract}
Solving Partial Differential Equations (PDEs) is ubiquitous in science and engineering.
Computational complexity and difficulty in writing numerical solvers has motivated the development of data-driven machine learning techniques to generate solutions quickly.
The recent rise in popularity of Large Language Models (LLMs) has enabled easy integration of text in multimodal machine learning models, allowing easy integration of additional system information such as boundary conditions and governing equations through text.
In this work, we explore using pretrained LLMs to integrate various amounts of known system information into PDE learning.
Using FactFormer as our testing backbone, we add a multimodal block to fuse numerical and textual information.
We compare sentence-level embeddings, word-level embeddings, and a standard tokenizer across 2D Heat, Burgers, Navier-Stokes, and Shallow-Water data sets.
These challenging benchmarks show that pretrained LLMs are able to utilize text descriptions of system information and enable accurate prediction using only initial conditions.
\end{abstract}

\section{Introduction}
\label{sec:intro}
Solving Partial Differential Equations (PDEs) is the cornerstone of many areas of science and engineering, from quantum mechanics to fluid dynamics.
While traditional numerical solvers often have rigorous error bounds, they are limited in scope, where different methods are required for different governing equations, and different regimes even for a single governing equation.
In the area of fluid dynamics, especially, solvers that are designed for Navier Stokes equations generally will not perform optimally in both the laminar and turbulent flow regimes.

Recently, machine learning methods have exploded in popularity to address these downsides in traditional numerical solvers.
The primary aim is to reduce time-to-solution and bypass expensive calculations.
Physics Informed Neural Networks (PINNs) \citep{RAISSI2019686} have been widely successful in small-scale systems, but are notoriously difficult to train \citep{krishnapriyan2021characterizingpossiblefailuremodes}.
More recently, neural operators \citep{li2021fourierneuraloperatorparametric, Lu2021, li2023transformerpartialdifferentialequations} have improved upon PINNs, showing promise for larger scale, general purpose surrogate models.
However, these neural operator models generally are purely data-driven, and do not use any additional, known, system information.

Additionally, owing to the success of large language models (LLMs) in many other domains, such as robotics \citep{kapoor2024logicallyconstrainedroboticstransformers, bartsch2024llmcraftroboticcraftingelastoplastic}, and design \citep{kumar2023mycrunchgptchatgptassistedframework, badagabettu2024query2cadgeneratingcadmodels, jadhav2024largelanguagemodelagent, jadhav2024llm3dprintlargelanguage}, some works have begun incorporating LLMs into the PDE surrogate model training pipeline.
Universal Physics Solver (UPS) \citep{shen2024upsefficientlybuildingfoundation} incorporates pretrained LLMs, but uses limited text descriptions that do not fully utilize LLM capabilities.
Unisolver \citep{zhou2024unisolverpdeconditionaltransformersuniversal} takes in a LaTeX description of the system as a prompt, but uses MLP encoders for additional system information such as boundary conditions, which does not fully explore the capabilities of the LLM.
ICON-LM \citep{yang2024finetunelanguagemodelsmultimodal} uses longer text descriptions, but trains the LLM to make numerical predictions from input data and captions, adding additional complexity to the model architecture.
Lastly, FMINT \citep{song2024fmintbridginghumandesigned} prompts an LLM to generate text descriptions, but only explores ODE systems.
Both FMINT and ICON-LM use softmax-based attention which may not scale well to larger systems \citep{cao2021choosetransformerfouriergalerkin}, as well.

\textbf{Contributions:} In this work, we explore differing levels of text description, fused with a mutlimodal block using pretrained LLMs.
Word-level embeddings from Llama \citep{llama3modelcard} are popular for Natural Language Processing tasks and are compared against sentence-level embeddings from SentenceTransformer \citep{reimers-2019-sentence-bert}, because global sentence information may be more useful in this case.
These are compared against a tokenizer from Llema \citep{azerbayev2023llemma} in order to evaluate how much the semantic knowledge of these pretrained LLMs affects downstream performance.
To that end, we introduce a cross-attention based multimodal block that integrates textual information into FactFormer \citep{li2023scalable}, given in figure \ref{fig:attention_mixer}.
We benchmark this multimodal approach on the Heat, Burgers, Navier Stokes, and Shallow Water equations to provide a wide variety of physical behavior.
Various boundary conditions, initial conditions, and operator coefficients are used, provide an additional challenge over existing benchmarks.
We describe this system information with text, and incorporate this as conditioning information into FactFormer.
\begin{figure}
    \centering
    \includegraphics[width=\linewidth]{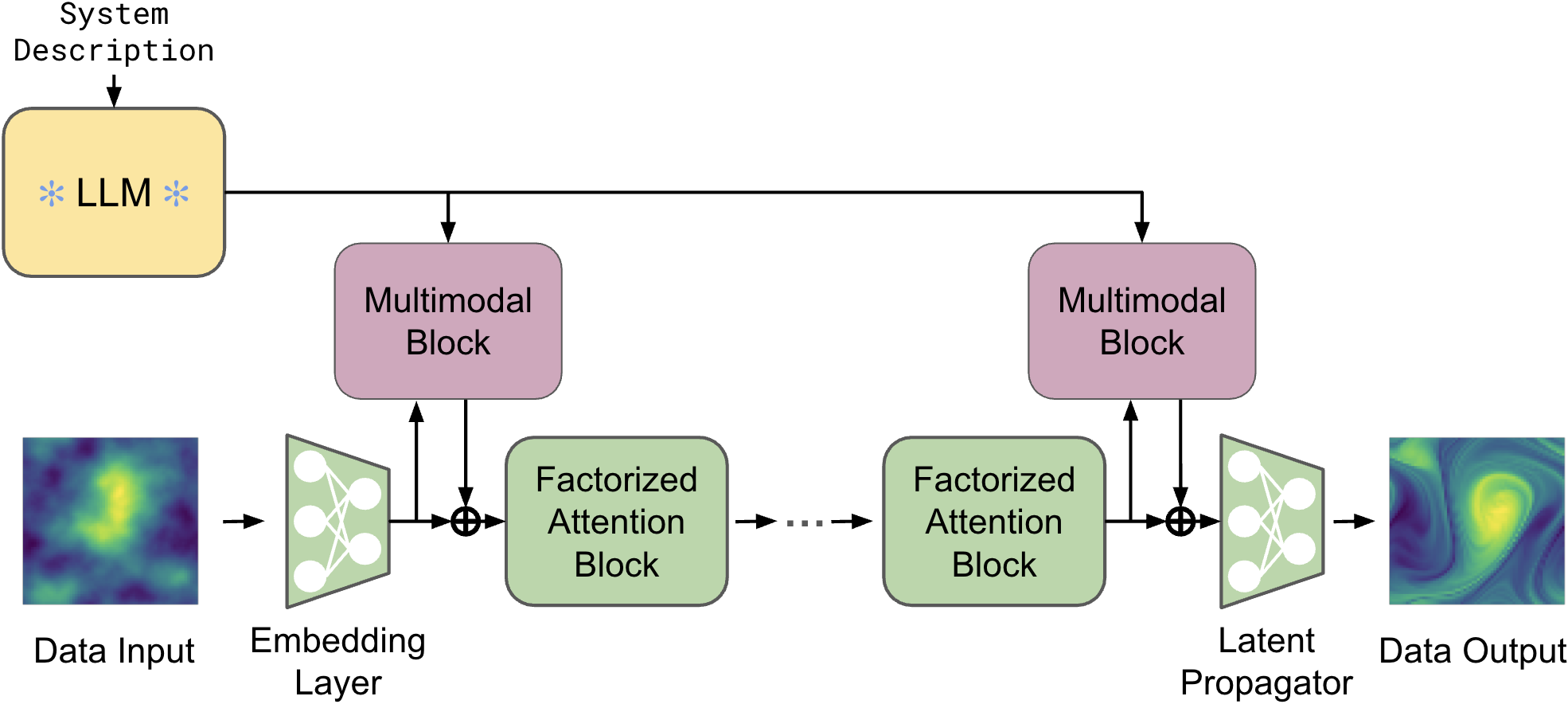}
    \caption{Multimodal FactFormer adds system information through a multimodal block.}
    \label{fig:multimodal}
\end{figure}

\section{Related Work}
\label{sec:related_work}

\textbf{Neural Solvers:} Neural solvers have been developed largely to counter the drawbacks of traditional numerical methods.
Physics Informed Neural Networks (PINNs) \citep{RAISSI2019686} incorporate governing equations through a soft constraint on the loss function.
PINNs have been shown to be effective in small-scale problems, but tend to be difficult to train \citep{WANG2022110768_pinnsdifficult, rathore2024challengestrainingpinnsloss}.
Neural operators \citep{JMLR:v24:21-1524kovachki} were developed and show improvement over PINNs on a large variety of PDE learning tasks.
Based on the universal operator approximation theorem, neural operators learn a functional that maps input functions to solution functions.
Neural operators such as Fourier Neural Operator (FNO) \citep{li2021fourierneuraloperatorparametric}, based on kernel integral blocks in Fourier space, DeepONet\cite{Lu2021}, based on parameterizing the input functions and an embedding of the points at which the functions are evaluated, and OFormer \citep{li2023transformerpartialdifferentialequations}, based on softmax-free attention.
Despite very promising results, neural operators are usually purely data-driven and do not use any system information outside of solution fields.

\textbf{Utilizing System Information} While these neural operators perform well, they are often purely data driven.
More recent works have incorporated various different aspects of the governing systems.
Takamoto et al. \yrcite{takamoto2023cape} incorporates operator coefficients through the CAPE module.
While coefficients are important in determining system behavior, system parameters such as boundary conditions, forcing terms, and geometry can play an equally important role, and cannot be easily incorporated through CAPE.
Further work incorporates governing equations into models.
Lorsung et al. \yrcite{pitt_Lorsung_2025} developed the PITT framework that uses a transformer-based architecture to process system information as text.
Additionally, PROSE \citep{liu2023prosepredictingoperatorssymbolic} poses a multi-objective task to both make a prediction and complete partially correct governing equation, and follow up work in PROSE-FD \citep{liu2024prosefdmultimodalpdefoundation_prosefd} is similar to PITT in incorporating symbolic representations as input.
These frameworks perform well, but rely on notational consistency between samples, and do not offer an easy integration of different geometries.
Lastly, Hao et al. \yrcite{hao2023gnotgeneralneuraloperator} introduced novel Heterogeneous Normalized Attention and Geometric Gating mechanisms for flexible GNOT model.
GNOT is able to incorporate different system information, such as coefficients and geometry.
This flexible architecture can incorporate many different system parameters, but requires additional implementation details for each additional modality, and may not be able to capture qualitative aspects of systems, such as flow regime.

\section{Data Generation}
We generate data from the Heat, Burgers, and Incompressible Navier Stokes equations, which are popular benchmarks for fluid dynamics surrogate models.
Additionally, we benchmark on the Shallow Water equations from PDEBench \citep{takamoto2023pdebenchextensivebenchmarkscientific}.
Data for the Heat and Burgers equations are generated using Py-PDE \citep{py-pde} due to ease of simulating different boundary conditions, and data is generated for Navier Stokes using code from Fourier Neural Operator \citep{li2021fourierneuraloperatorparametric}.
A diverse data set is generated in order to create a more challenging setup than existing benchmarks, that often have the same operator coefficients and boundary conditions for all samples, only varying the initial conditions.
In our case, we use different initial conditions, operator coefficients, and boundary conditions.
While existing data sets offer distinct challenges with regards to governing equations, they do not offer a lot of data diversity with regard to system parameters, which LLMs are well-suited to handle.
PDEBench, for example, has multiple 2D data sets that represent a variety of physical processes.
However, only eight different operator coefficient combinations are used for the 2D Compressible Navier Stokes equations, all with the same boundary conditions.
In order to fully utilize the capabilities of pretrained LLMs, as well as present a more challenging benchmark, we vary boundary conditions and operator coefficients for the Heat and Burgers equations in 2D, and vary viscosity and forcing term amplitude for the Incompressible Navier-Stokes equations in 2D.

\begin{figure}
    \centering
    \includegraphics[width=0.9\linewidth]{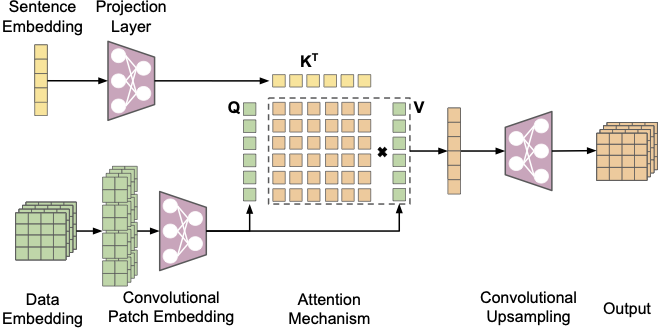}
    \caption{Cross-Attention is used to integrate sentence embeddings with data embeddings.}
    \label{fig:attention_mixer}
\end{figure}

\subsection{Heat Equation}
The heat equation models a diffusive process and is given below in equation \ref{eq:heat}, where we are predicting the temperature distribution at each time step.
\begin{equation}
    u_{t} = \beta\nabla^2 u
    \label{eq:heat}
\end{equation}
We generate data on a simulation cell given by: $\Omega = \left[-0.5, 0.5\right]^2$ on a 64x64 grid with boundary conditions sampled from $\partial\Omega \in \left\{Neumann, Dirichlet, Periodic\right\}$.
The values $v$ for the Neumann and Dirichlet boundary conditions are sampled uniformly: $v \in \mathcal{U}(-0.1, 0.1)$.
All four walls have the same boundary condition type and value for a given simulation.
Initial conditions are given by $f(x,y) = \exp\left(100(x+y)^2\right)$
Lastly, our diffusion coefficient was sampled randomly according to $\beta \in\mathcal{U}\left(0.01, 0.001\right)$.
We simulated each trajectory for 1 second on a 64x64 grid that was evenly downsampled to 100 frames, for 101 total snapshots of data including initial conditions.
These distribution of diffusion coefficients was chosen so the diffusive dynamics behaved on approximately the same scale as the advection dynamics of Burgers Equation given below.

\subsection{Burgers' Equation}
Burgers equation models shock formation in fluid waves, given below in equation \ref{eq:burgers}, where we are predicting the height, $u$ at each time step.
\begin{equation}
    u_{t} = \beta\nabla u - \alpha u\cdot \nabla u
    \label{eq:burgers}
\end{equation}
System specifications for the Heat equation are identical to Heat equation, given above, with the exception of the advection term.
Our advection coefficient is sampled according to $\alpha_x, \alpha_y \in \mathcal{U}\left(-1.0, 1.0\right)$.

\subsection{Incompressible Navier-Stokes Equations}
Third, we generate data from the Incompressible Navier-Stokes equations in vorticity form, given below in equation \ref{eq:navier_stokes}.
\begin{equation}
\begin{split}
    \partial_t w(x,t) + u(x,t)\cdot\nabla w(x,t) &= \nu \Delta w(x,t) + f(x) \\
    f(x) = A(\sin\left(2\pi\left(x_1 + x_2\right)\right) +& \cos\left(2\pi\left(x_1 + x_2\right)\right)) \\
    \nabla \cdot u(x,t) &= 0 \\
    w(x,0) &= w_0(x)
\end{split}
\label{eq:navier_stokes}
\end{equation}
The Navier-Stokes data generated here follows the setup from \cite{pitt_Lorsung_2025} with different viscosities and forcing term amplitudes.
Our simulation cell is given by: $\Omega = \left[0, 1\right]^2$ with periodic boundary conditions.
Our viscosity is sampled according to 

$\nu\in\left\{10^{-8}, 5\cdot10^{-8}, 10^{-7}, 10^{-6}, 5\cdot 10^{-6}, 10^{-5}, 5\cdot 10^{-5}\right\}$, and our forcing term amplitude is sampled according to $A \in \left\{0.001, 0.005, 0.01, 0.05, 0.1\right\}$.
The initial condition is given by a Gaussian random field and out simulation is run for 10 seconds.
Data generation is done on a $256\times256$ grid that is evenly downsampled to a 64x64 grid for this work, and 100 evenly sampled timesteps with initial condition, for 101 total snapshots of data.

\subsection{Shallow-Water Equations}
Lastly, we take the Shallow-Water data set from PDEBench  \citep{takamoto2023pdebenchextensivebenchmarkscientific}.
In this case, we are predicting the water height $h$ at each time , where $u$ is x-velocity, $v$ is y-velocity, $g_r$ is gravity, and $b$ is the bathymetry.
\begin{equation}
\begin{split}
    \partial_t h + \partial_x h u + \partial_y h v &= 0 \\
    \partial_t h u + \partial_x\left(u^2h + \frac{1}{2}g_rh^2 \right) + \partial_y u v h &= -g_rh\partial_xb \\
    \partial_t h v + \partial_y\left(v^2h + \frac{1}{2}g_rh^2 \right) + \partial_x u v h &= -g_rh\partial_yb \\
\end{split}
    \label{eq:shallow_water}
\end{equation}
Our simulation cell is: $\Omega = [-2.5, 2.5]^2$ on a 64x64 grid, with Neumann boundary conditions, with 0 gradient on the boundary.


\section{Methods}
The multimodal appraoch developed here uses full sentence descriptions of systems from our data sets, given in section \ref{sec:system_description}.
The cross-attention based multimodal block is uses both FactFormer embeddings as well as LLM embeddings and is described in section \ref{sec:multimodal}.

\subsection{System Descriptions}
\label{sec:system_description}
To explore how well our LLMs are able to incorporate different amounts of text information, we describe each of the Heat, Burgers, and Navier-Stokes equations with varying levels of detail.
We will build a complete sentence description here, with all possible sentence combinations given in appendix \ref{app:sentences}.
At a base level, we can describe the basic properties of each governing equation, as well as identifying the equation.
For Burgers equation, we generally have stronger advection forces, so we describe it as:

    \small\texttt{Burgers equation models a conservative system that can develop shock wave discontinuities. Burgers equation is a second order partial differential equation.}

All of our basic system descriptions come from Wikipedia entries for their respective equations  \citep{wiki:Burgers'_equation, wiki:Heat_equation, wiki:Navier–Stokes_equations, wiki:Shallow_water_equations}.
Next, boundary condition information is added, in this case Neumann boundary conditions:

    \small\texttt{This system has Neumann boundary conditions. Neumann boundary conditions have a constant gradient. In this case we have a gradient of $\partial u_{neumann}$ on the boundary.}

Third, we can add operator coefficient information:

    \small\texttt{In this case, the advection term has a coefficient of $\alpha_x$ in the x direction, $\alpha_y$ in the y direction, and the diffusion term has a coefficient of $\beta$.}

Lastly, we can add qualitative information. The aim of this is to capture details about the system that are intuitive to practitioners, but difficult to encode mathematically. In this case, we have an advection dominated system:

    \small\texttt{This system is advection dominated and does not behave similarly to heat equation.  The predicted state should develop shocks.}

Our complete sentence description is passed into a pretrained LLM that is used to generate embeddings.
These embeddings are then used as conditioning information for our model output.

\subsection{Multimodal PDE Learning}
\label{sec:multimodal}
The backbone of our multimodal surrogate model is the FactFormer \citep{li2023scalable} and our framework is given in figure \ref{fig:multimodal}.
FactFormer was chosen because it provides a fast and accurate benchmark model.
We add our system description as conditioning information both before and after factorized attention blocks.
FactFormer is a neural operator that learns a functional $\mathcal{G}_{\theta}$ that maps from input function space $\mathcal{A}$ to solution function space $\mathcal{U}$ as $\mathcal{G}_{\theta}: \mathcal{A} \to \mathcal{U}$, with parameters $\theta$.
In our case, we are specifically learning the functional conditioned on our system information $\mathbf{s}$.
For a given input function $a$ evaluated at points $\mathbf{x}$, we are learning the the operator given in equation \ref{eq:operator_learning}.
That is, our network learns to make predictions for our solution function $u$ also evaluated at points $\mathbf{x}$.
\begin{equation}
    G(u)(\mathbf{x}) = \mathcal{G}_{\theta}(a)(\mathbf{x}, \mathbf{s})
    \label{eq:operator_learning}
\end{equation}

Sentences are first passed through an LLM to generate an embedding.
The embedding is combined with the data through cross-attention, seen in figure \ref{fig:attention_mixer}, where our sentence embedding is the queries and the data embedding is the keys and values.
The conditioned embedding is then added back to our data embedding, given below in equation \ref{eq:multimodal}.
Our data embedding is embedding with convolutional patch embedding, mapping our data to a lower dimension as $f_{data}: \mathcal{R}^{b\times h \times w \times h_{FF}} \to \mathcal{R}^{b \times p \times h_{FF}}$, where $b$ is our batch size, $h_{FF}$ is our FactFormer embedding dimension, and $p$ is the number of patches, which is defined by kernel size and convolutional stride.
Our sentence embeddings are projected to higher dimensional space to match our data embedding dimension, $f_{sentence}:\mathcal{R}^{b\times h_{LLM}} \to \mathcal{R}^{b\times p\times 1} \to \mathcal{R}^{b \times p \times h_{FF}}$ through two successive MLPs, where $h_{LLM}$ is the LLM output dimension.
Once we have the data embedding $\mathbf{z}_{data} = f_{data}(\mathbf{a})$ and our sentence embedding $\mathbf{z}_{sentence} = f_{sentence}(\mathbf{s})$, for data sample $\mathbf{a}$ and sentence description $\mathbf{s}$, we calculate the multimodal embedding $\mathbf{z}_{multimodal}$ using multihead attention \citep{NIPS2017_3f5ee243attentionisallyouneed} given below in equation \ref{eq:multimodal}.
\begin{equation}
    \begin{split}
    \mathbf{z}_{multimodal} &= \text{Concat}\left(\text{head}_1,\ldots, \text{head}_h\right)\mathbf{W}^O\\
    \text{and head}_i &= \text{softmax}\left(\frac{\mathbf{z}_{sentence}\mathbf{W}_{i}^{Q} \left(\mathbf{z}_{data}\mathbf{W}_{i}^{K}\right)^T}{\sqrt{d_k}}\right)\mathbf{z}_{data}\mathbf{W}_{i}^{V}
    \end{split}
    \label{eq:multimodal}
\end{equation}
$\mathbf{z}_{multimodal}$ is then returned from our multimodal block and used in the FactFormer architecture.
We use cross-attention here instead of self-attention so that our sentence embeddings can be used to learn useful context for our data embeddings.
After cross-attention, our multimodal embedding is upsampled back to the data embedding dimension using deconvolution \citep{5539957convtranspose} $f_{deconv}: \mathcal{R}^{b \times p \times h_{FF}} \to \mathcal{R}^{b\times h \times w \times h_{FF}}$ given in equation \ref{eq:recombine}:
\begin{equation}
    \mathbf{z}_{data} = \mathbf{z}_{data} + f_{deconv}\left(\mathbf{z}_{multimodal}\right)
    \label{eq:recombine}
\end{equation}

\subsection{Text Processing}
\label{sec:text_processing}
This multimodal block is agnostic to both the data-driven backbone as well as the LLM.
We compare \texttt{Llama 3.1 8B} \citep{llama3modelcard}, and \texttt{all-MiniLM-L6-v2} from the Sentence-Transformer (ST) package \citep{reimers-2019-sentence-bert} as our pretrained LLMs, as well as the tokenizer from Llema \citep{azerbayev2023llemma}.
$\texttt{Llama 3.1 8B}$ was used because of its good performance across a wide variety of benchmarks, as well as its small size allowing us to use only a single GPU to generate sentence embeddings.
The word embeddings are averaged for each sentence to provide a single embedding that can be used in our projection layer.
$\texttt{all-mpnet-base-v2}$ was used because it is designed to generate embeddings from sentences that are useful for tasks like sentence similarity.
This allows downstream tasks to utilize embeddings that utilize the entire sentence, similar to human reasoning at the sentence level.
Lastly, our tokenizer-only variant is used to provide insight in how much semantic knowledge our models are able to utilize from our pretrained LLMs, because a tokenizer alone has none.
In all of our experiments, the LLM is frozen and a projection head is trained.
This significantly improves training time by allowing us to generate the sentence embeddings before training, and avoiding expensive gradient computations for our LLMs.

\section{Results}
\label{sec:results}
We benchmark our multimodal model against its baseline variant on a number of challenging tasks.
Our data vary the distribution of initial conditions, operator coefficients, and boundary conditions, which provides a much more challenging setting than many existing benchmarks.
In each experiment, the combined data set is the Heat, Burgers, and Navier-Stokes data sets, where Shallow Water is used solely for fine-tuning, with a 80-10-10 train-validation-test split.
In our transfer learning experiments, we first train on this combined data set, then finetune on each individual data set.
Transfer learning was done because \citet{zhou2024strategiespretrainingneuraloperators} found it to be an effective pretraining strategy.
1,000 samples per equation are used for each data set.
The effect of text information is evaluated on next-step prediction, autoregressive rollout, and fixed-future prediction tasks.
We use only the initial condition as data input in all of our experiments.
Using only a single step of input here presents the additional challenge that it is difficult to infer boundary conditions and impossible to infer operator coefficients from only a single frame of data.
Relative $L^2$ Error \citep{li2021fourierneuraloperatorparametric} is used for both training and next-step evaluation in all of our experiments.
Autoregressive rollout error is reported as Mean Squared Error.
Experiments use various combinations of boundary condition (B), coefficient (C) and qualitative (Q) information.
Each reported result is the mean and standard deviation across three random seeds.
Results for our combined data sets, as well as tabular values for training are given in appendix \ref{app:additional_results}.
Training details are given in appendix \ref{app:training_setup}.

\subsection{Next-Step Prediction}
\label{sec:next_step_pred}
\begin{figure}[h]
    \centering
    \includegraphics[width=\linewidth]{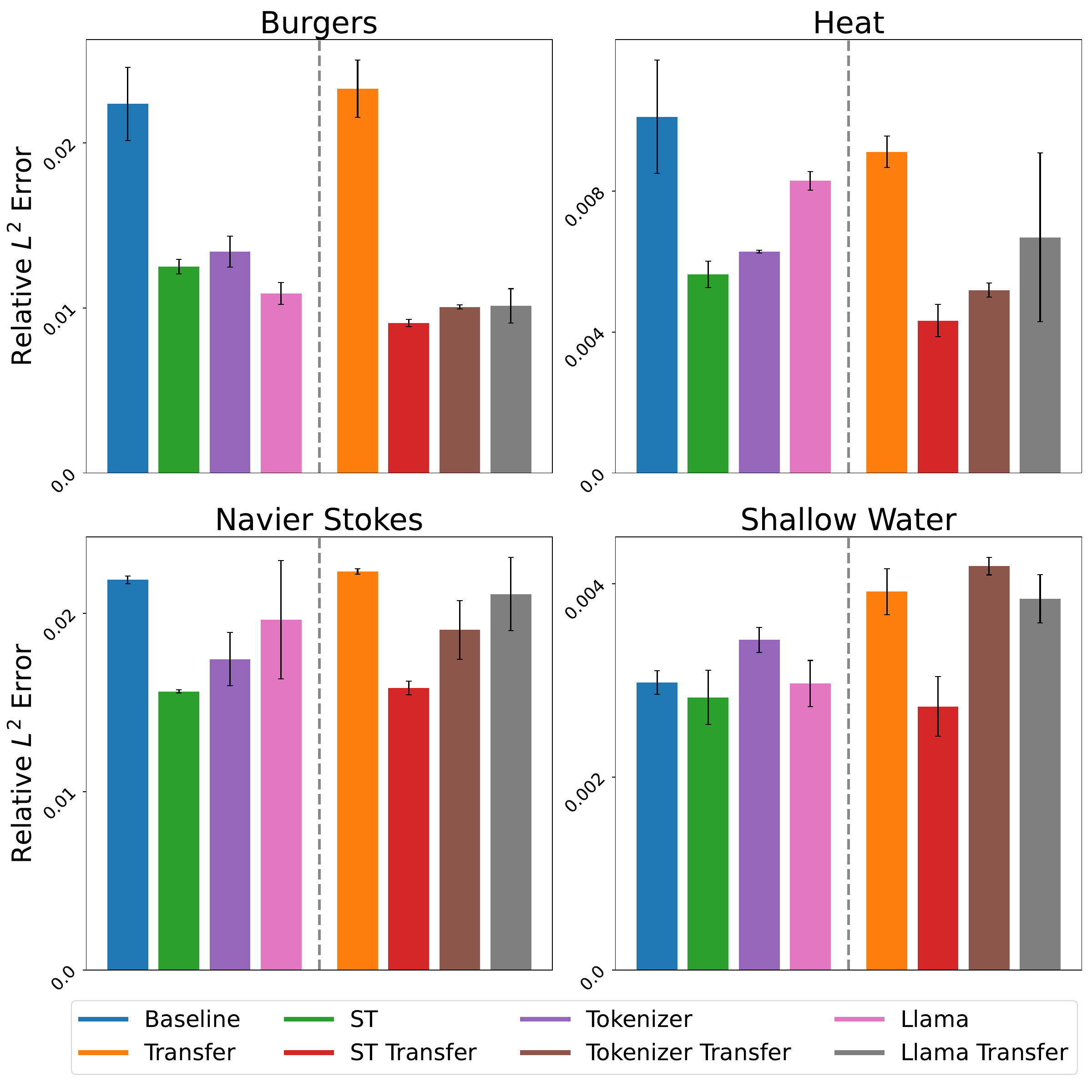}
    \caption{Comparison of next-step prediction relative $L^2$ error for baseline FactFormer, FactFormer + ST, FactFormer + Tokenizer, and Factformer + Llama with and without transfer learning.}
    \label{fig:pred_comparison}
\end{figure}

%
%
First, we evaluate next-step predictive accuracy, where we take one frame of data and use it to predict the next across our entire temporal window.
We see in figure \ref{fig:pred_comparison} that our multimodal FactFormer significantly outperforms baseline FactFormer for each of our multimodal approaches.
In this case, SentenceTransformer either has best performance, or is close to best performance across all of our data sets with both standard training and transfer learning.
Both our tokenizer and Llama variants offer improvement over baseline as well.
Because our Shallow Water data set does not have varying coefficients, boundary conditions, or qualitative behavior, we see comparable performance between baseline and multimodal models.
This is corroborated in our ablation study in section \ref{sec:ff_results}.
When excluding the Shallow-Water results, our SentenceTransformer variant has an average reduction in error of 39.0\% and 47.6\% for standard training and transfer learning, respectively, when compared to our baseline method.

\subsection{Autoregressive Rollout}
\begin{figure}
    \centering
    \includegraphics[width=\linewidth]{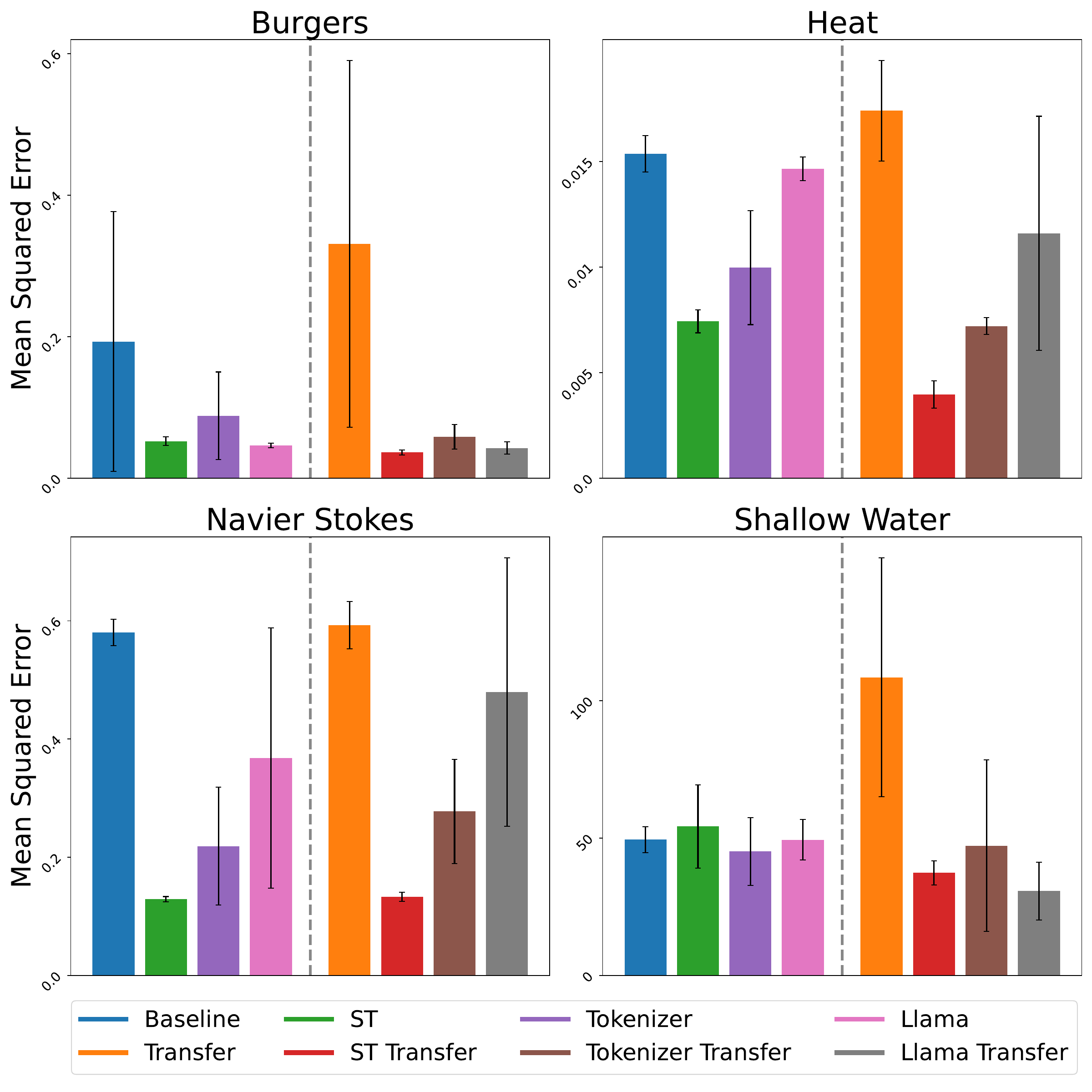}
    \caption{Comparison of accumulated mean squared error for baseline FactFormer, FactFormer + ST, FactFormer + Tokenizer, and baseline + Llama with and without transfer learning.}
    \label{fig:accumulated_error}
\end{figure}

%
%
Second, we perform autoregressive rollout starting from our initial condition for 40 steps.
This evaluates how well our model is able to temporally extrapolate beyond our training horizon.
In this case, we use the models trained in section \ref{sec:next_step_pred} with no additional fine-tuning.
Accumulated rollout error is presented in figure \ref{fig:accumulated_error}.
Our autoregressive rollout error results follow closely from the next-step prediction results.
We see that our SentenceTransformer variant has lowest error for our Heat, Burgers, and Navier Stokes data sets for both standard training and transfer learning.
On our Shallow Water data sets, we note comparable performance for all models for the reasons noted in section \ref{sec:next_step_pred}.
Similarly, when excluding the Shallow Water results, the SentenceTransformer variant has an average reduction in accumulated error of 67.4\% and 81.3\% for standard training and transfer learning, respectively.

We note that accumulated autoregressive rollout error for our Shallow Water benchmark is much higher for our transfer learning experiments than the standard training.
This instability been noted before in \cite{pitt_Lorsung_2025} when training errors are very low.
In our case, training error for the Shallow Water data set is approximately an order of magnitude lower than other data sets.

%

\subsection{Fixed-Future}
\label{sec:ff_results}
\begin{figure}[h]
    \centering
    \includegraphics[width=\linewidth]{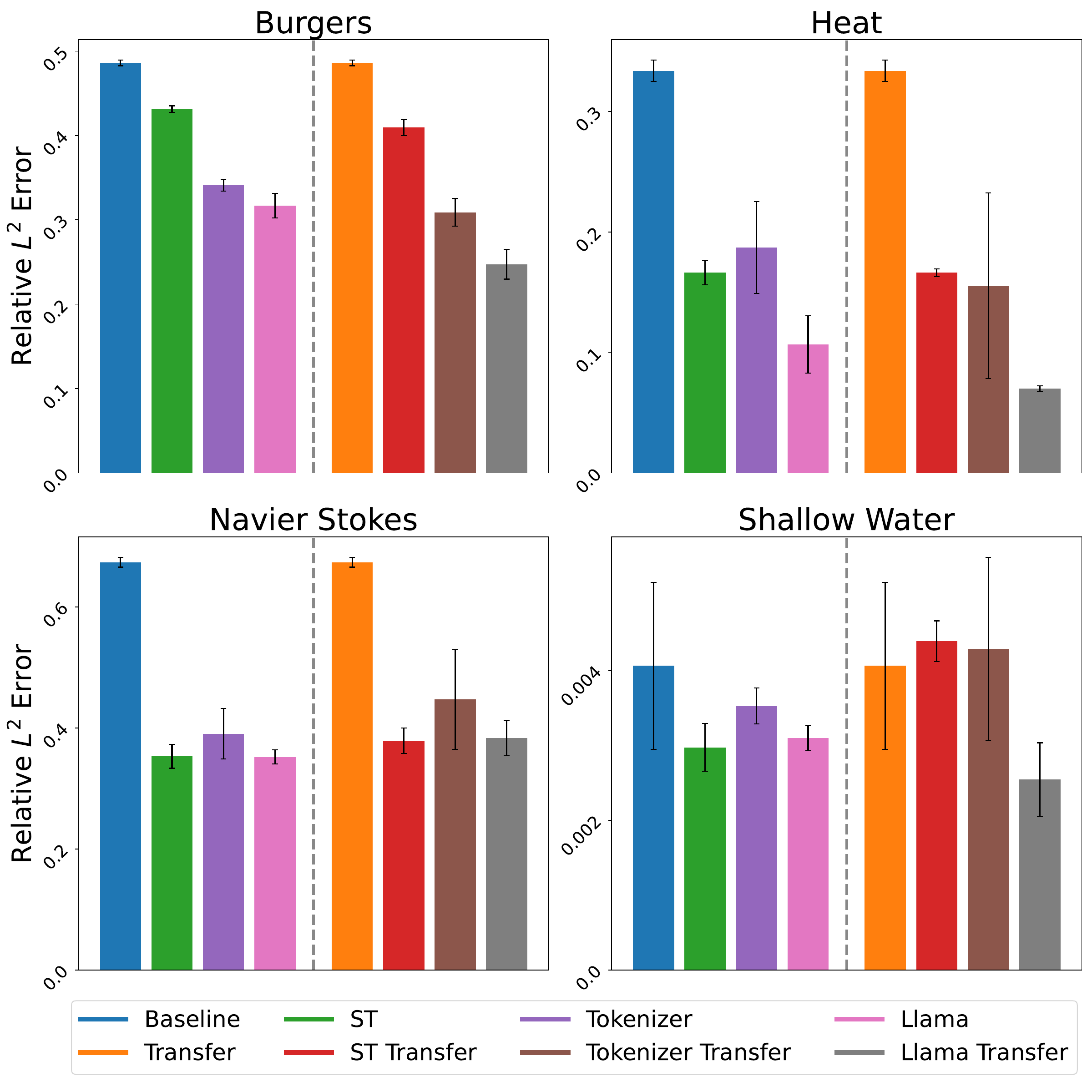}
    \caption{Comparison of fixed-future prediction relative $L^2$ error for baseline FactFormer, FactFormer + ST, FactFormer + Tokenizer, and FactFormer + Llama with and without transfer learning.}
    \label{fig:ff_results}
\end{figure}

%
%
In this case, we use the initial condition to predict the final simulation state for each data set.
Results are given in figure \ref{fig:ff_results}.
We see clear improvement over baseline in the Heat, Burgers, and Navier-Stokes data sets for each of our multimodal variants, with marginal improvement over Shallow Water for the same reasons given in section \ref{sec:next_step_pred}.
Interestingly, we see that our Llama variant has greatest improvement over baseline, in contrast to the next-step and autoregressive rollout experiments.
Our Llama variant has an average reduction in error of 50.2\% and 57.1\% when excluding Shallow Water for standard training and transfer learning, respectively.

\subsection{Ablation Study}
Lastly, we perform an ablation study based on text information to determine which components of our text description are most helpful during the learning process using our SentenceTransformer variant.
Results are given in figure \ref{fig:ablation_ff}.
An additional ablation study with our Llama variant is given in appendix \ref{app:ablation}.
The same experimental setup is used as in section \ref{sec:ff_results}.
We see that performance correlates strongly with relevant sentence information.
Only the viscosity and forcing term amplitude vary in our Navier Stokes data set, with periodic boundary conditions in all samples.
We see that adding boundary condition information has no impact on performance.
Similarly, in the Shallow Water data set, we have the same boundary conditions and flow conditions for each sample, and so there is a very weak correlation with additional system information.
There is also a weak performance improvement with the addition of qualitative information.
Qualitative information acts as an imperfect proxy for coefficient information for the Heat, Burgers, and Navier Stokes data sets.
This leads to a small improvement in performance when it is present without coefficient information.
However, when coefficient information is provided, we generally do not see much additional improvement in performance.
With our Heat data set, we see improvement as more information is added, with full system information having best performance.
However, with Burgers equation, we see weaker correlation, suggesting that the LLM is unable to learn more complex systems and distribution of values as well as the simpler heat equation, or discrete Navier Stokes coefficients.

\begin{figure}[h]
    \centering
    \includegraphics[width=\linewidth]{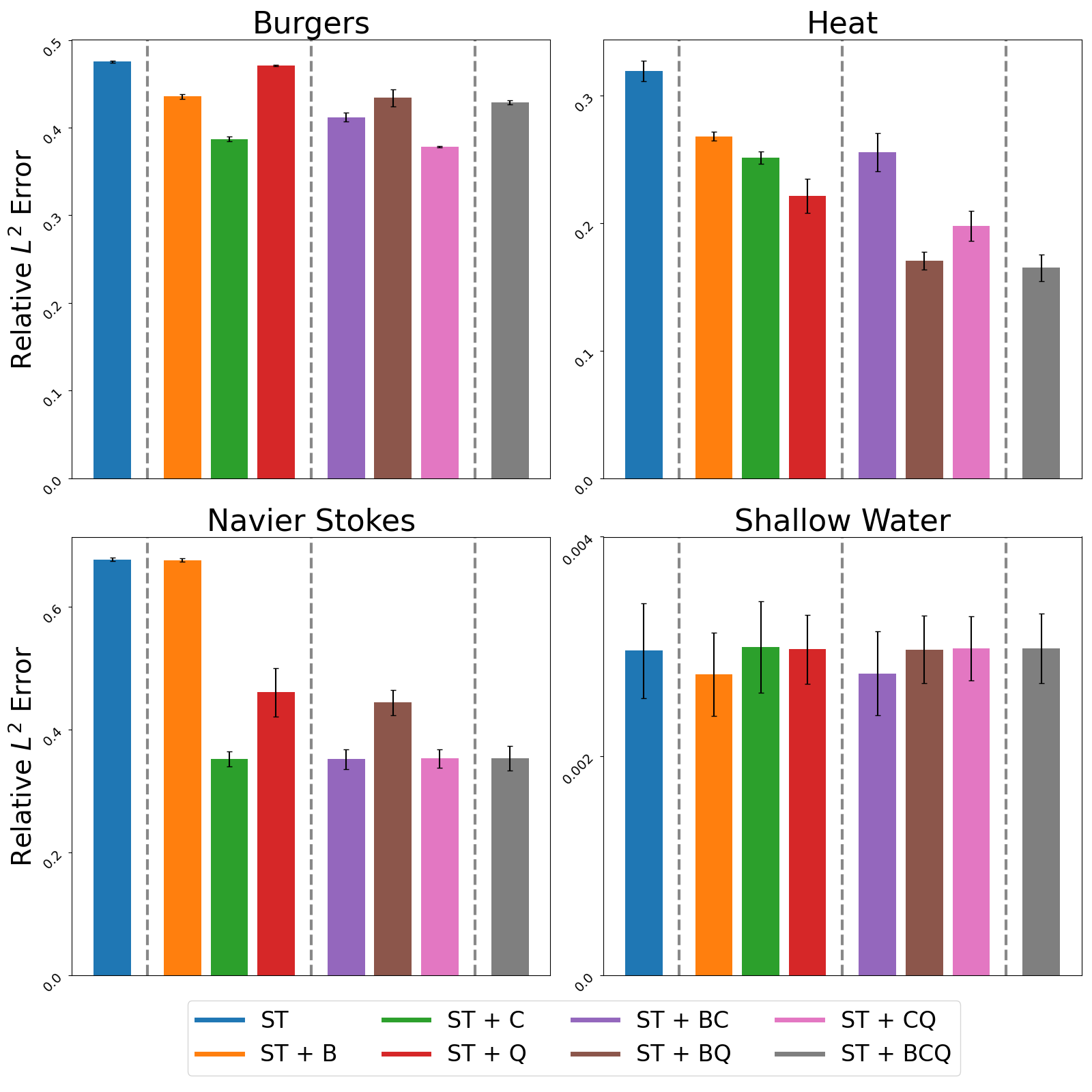}
    \caption{Comparison of fixed-future prediction relative $L^2$ error for FactFormer + ST with varying levels of sentence description, trained with standard training.}
    \label{fig:ablation_ff}
\end{figure}

\section{Discussion}
The aim of our multimodal approach is to incorporate known system information through sentence descriptions, rather than complex data conditioning strategies.
We see for our next-step prediction and autoregressive rollout tasks the SentenceTransformer variant tends to perform best, but on the fixed-future task Llama performs best, while the tokenizer variant is between the two in performance.
Although Llama is larger than SentenceTransformer, these results do not show a clear trend that Llama's capacity is being utilized for this task.
Additionally, our tokenizer variant having comparable performance to both pretrained LLMs suggests the semantic knowledge of these LLMs does not affect performance.
This is also seen in the ablation study results, where qualitative information has a weak effect on performance, and almost no effect when coefficient information is present.

\begin{figure}[h]
    \centering
    \includegraphics[width=\linewidth]{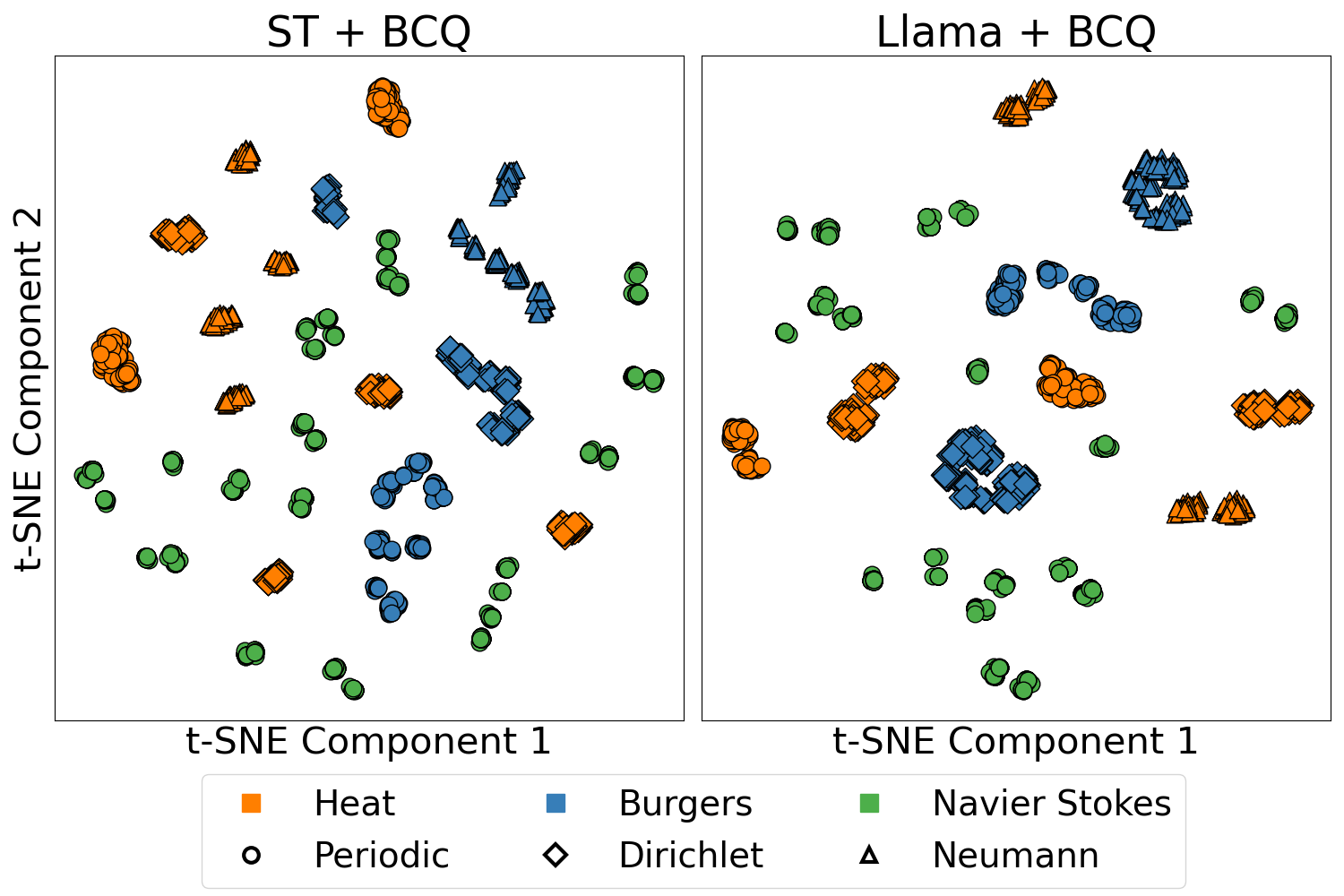}
    \caption{t-SNE embeddings for sentence-level embeddings generated by SentenceTransformer and word-level embeddings generated by Llama for equation descriptions with boundary conditions, operator coefficients, and qualitative information.}
    \label{fig:none_vs_all_tsne}
\end{figure}

We can visualize the embeddings from SentenceTransformer and Llama for each level of text description as well in our combined dataset.
In figure \ref{fig:none_vs_all_tsne}, we have t-SNE embeddings for sentences with the full text description, that is with boundary condition, operator coefficient, and qualitative information.
t-SNE plots for all other combinations of information are given in appendix \ref{app:tsne}.
For both SentenceTransformer and Llama, we have clusters corresponding to combinations of boundary conditions and governing equation
This is despite that we average over sequence length to obtain Llama text embeddings.
While we see clear clusters for equations and boundary conditions, we do not see mixed clusters seen in PICL \citep{lorsung2024piclphysicsinformedcontrastive}.
These clusters do not match intuition that diffusion dominated Burgers equation behaves similarly to the Heat equation, suggesting that these pretrained LLMs lack semantic understanding of PDE systems.
t-SNE plots were generated with TSNE-CUDA \cite{chan2019gpu}.

\section{Conclusion}
We introduced a multimodal approach to PDE surrogate modeling that utilizes LLMs and sentence descriptions of our systems that significantly improves performance for a variety of tasks.
Our multimodal approach allows us to easily incorporate system information that both captures quantitative and qualitative aspects of our governing equations.
Analysis of our sentence embeddings shows that we are able to capture increasing amounts of underlying structure in our data by simply adding more information through text, rather than introducing additional mathematical approaches, such as constrained loss functions that are popular with PINNs.
However, marginal improvement when using qualitative information, and comparable performance to a simple tokenizer suggests that the semantic understanding of LLMs is not being fully utilized in this case.

While this direction is promising, performance could be further improved with tuning LLMs directly, rather than just training an embedding layer that uses LLM output.
PDEs are defined rigorously and precisely, and so training LLMs specifically for PDE applications
may also allow the semantic understanding of LLMs to be used in similar settings.
We leave this to future work.
Additionally, pretraining strategies have shown success in various LLM applications from natural language processing to PDE surrogate modeling.
While this does introduce significant computational overhead, it may improve performance and is also a potential future direction to take.
Lastly, we only benchmarked the FactFormer, but this framework and multimodal block is not specific to FactFormer, and may prove useful for improving other popular PDE surrogate models such as FNO and DeepONet.

\section*{Impact Statement}
This paper presents work whose goal is to advance the field of 
Machine Learning. There are many potential societal consequences 
of our work, none which we feel must be specifically highlighted here.


\bibliography{the}
\bibliographystyle{icml2025}

\newpage
\appendix
\onecolumn
\section{Sentence Descriptions}
\label{app:sentences}
Basic information:\\

\begin{tabularx}{\textwidth}{lX}
    Heat: &  \small\texttt{The Heat equation models how a quantity such as heat diffuses through a given region. The Heat equation is a linear parabolic partial differential equation.}\\
    \\
    Burgers: & \small\texttt{Burgers equation models a conservative system that can develop shock wave discontinuities. Burgers equation is a second order partial differential equation.}\\
    \\
    Navier Stokes: & \small\texttt{The incompressible Navier Stokes equations describe the motion of a viscous fluid with constant density. We are predicting the vorticity field, which describes the local spinning motion of the fluid.}\\
    \\
    Shallow Water: & \small\texttt{The Shallow-Water equations are a set of hyperbolic partial differential equations that describe the flow below a pressure surface in a fluid.}\\
\end{tabularx}
\\

Coefficient Information:\\

\begin{tabularx}{\textwidth}{lX}
    Heat: & \small\texttt{In this case, the diffusion term has a coefficient of $\beta$.}\\
    \\
    Burgers: & \small\texttt{In this case, the advection term has a coefficient of $\alpha_x$ in the x direction, $\alpha_y$ in the y direction, and the diffusion term has a coefficient of $\beta$.}\\
    \\
    Navier-Stokes: & \small\texttt{In this case, the viscosity is $\nu$. This system is driven by a forcing term of the form f(x,y) = A*(sin(2*pi*(x+y)) + cos(2*pi*(x+y))) with amplitude A=$A$.}\\
    \\
    Shallow-Water: & \small N/A \\
\end{tabularx}
\\
Boundary Condition Information:\\

\begin{tabularx}{\textwidth}{llX}
    Heat: & Periodic & \small\texttt{This system has periodic boundary conditions. The simulation space is a torus.} \\
    & Neumann: & \texttt{This system has Neumann boundary conditions. Neumann boundary conditions have a constant gradient. In this case we have a gradient of {} on the boundary.} \\
    & Dirichlet: & \texttt{This system has Dirichlet boundary conditions. Dirichlet boundary conditions have a constant value. In this case we have a value of {} on the boundary.}\\
    \\
    Burgers: & Periodic & \small\texttt{This system has periodic boundary conditions. The simulation space is a torus.} \\
    & Neumann: & \texttt{This system has Neumann boundary conditions. Neumann boundary conditions have a constant gradient. In this case we have a gradient of {} on the boundary.} \\
    & Dirichlet: & \texttt{This system has Dirichlet boundary conditions. Dirichlet boundary conditions have a constant value. In this case we have a value of {} on the boundary.}\\
    \\
    Navier-Stokes: & Periodic: & \small\texttt{This system has periodic boundary conditions. The simulation cell is a torus.} \\
    \\
    Shallow-Water: & Neumann & \small\texttt{This system has homogeneous Neumann boundary conditions with a derivative of 0 at the boundary.}\\
\end{tabularx}\\
\clearpage

Qualitative Information:\\

\begin{tabularx}{\textwidth}{llX}
    Heat: & $\beta > 0.005$ & \small\texttt{This system is strongly diffusive. The predicted state should look smoother than the inputs.} \\
    & $\beta \leq 0.005$ & \small\texttt{This system is weakly diffusive. The predicted state should looke smoother than the inputs.}
    \\
    Burgers: & $\frac{\left\lVert \mathbf{\alpha} \right\rVert_2}{\beta} > 200$ & \small\texttt{This system is advection dominated and does not behave similarly to heat equation.  The predicted state should develop shocks.}\\
    & $\frac{\left\lVert \mathbf{\alpha} \right\rVert_2}{\beta} \leq 200$ & \small\texttt{This system is diffusion dominated and does behave similarly to heat equation. The predicted state should look smoother than the inputs.}\\
    \\
    Navier-Stokes: & $\nu \geq 1E-6$ & \small\texttt{This system has high viscosity and will not develop small scale structure.}\\
    & $1E-6 > \nu \geq 1E-8$ & \small\texttt{This sytem has moderate viscosity and will have some small scale structure.}\\
    & $1E-8 > \nu$ & \small\texttt{This system has low viscosity and will have chaotic evolution with small scale structure.} \\
    & $A \geq 7E-4$ & \small\texttt{This system has a strong forcing term and evolution will be heavily influenced by it.} \\
    & $7E-4 > A \geq 3E-4$ & \small\texttt{This system has a moderate forcing term and evolution will be moderately influenced by it.} \\
    & $3E-4 > A$ & \small\texttt{This system has a weak forcing term and evolvution will be weakly influenced by it.} \\
    \\
    Shallow-Water: & & \small\texttt{This system simulates a radial dam break. Waves propagate outward in a circular pattern.}\\
\end{tabularx}
\newpage

\section{Training Setup}
\label{app:training_setup}
\subsection{Model Architecture}
Model architecture was kept the same for the multimodal FactFormer variants for each experiment.
We note that the hidden dimension controls both the FactFormer embedding dimension as well as the LLM projection head embedding dimension.
This is done so we can use our multimodal cross-attention block without any additional data projections.
Our convolutional patch embedding used a kernel size of 8, and our convolutional upsampling layer used a ConvTranspose layer with kernel size of 8 with a stride size of 4, followed by two fully connected layers with widths of 128, and GELU activation.
Additionally, our sentence embedding projection layer used a single hidden layer of size [384/4096/4096, no. patches] project the embedding dimension of SentenceTransformer, Llama, and Llema, respectively, down to match the number of patches from our patch embedding.
The embeddings are then upsamples our channel dimension of 1 to match our hidden dimension using a single layer.

\subsection{Next-Step}
\begin{table}[h]
    \caption{FactFormer architecture hyperparameters.}
    \resizebox{\linewidth}{!}{
    \begin{tabular}{c|cccccc}
         Model & Depth & Hidden Dim & Head Dim & Heads & Kernel Multiplier & Latent Multiplier \\
         \hline
         FactFormer & 1 & 32 & 4 & 4 & 2 & 2 \\
         Multimodal FactFormer & 1 & 32 & 1 & 1 & 2 & 2 \\
    \end{tabular}}
    \label{tab:next_step_architecture}
\end{table}
\subsection{Fixed-Future}
\begin{table}[h]
    \caption{FactFormer architecture hyperparameters.}
    \resizebox{\linewidth}{!}{
    \begin{tabular}{c|cccccc}
         Model & Depth & Hidden Dim & Head Dim & Heads & Kernel Multiplier & Latent Multiplier \\
         \hline
         FactFormer & 1 & 64 & 4 & 4 & 2 & 2 \\
         Multimodal FactFormer & 1 & 32 & 4 & 4 & 2 & 2 \\
    \end{tabular}}
    \label{tab:fixed_future_architecture}
\end{table}

\subsection{Hyperparameters}
\label{app:hyperparams}
Training hyperparameters such as batch size, learning rate, and weight decay were hand tuned for each experiment to reduce error.
All models were trained with the Adam optimizer \citep{kingma2017adammethodstochasticoptimization}.

\newpage

\section{Additional Training Results}
\label{app:additional_results}
Results for our combined data set, numerical values for each experiment, and the ablation study with Llama as the LLM backbone are presented here.
For all results, $\textbf{bold}$ indicates the lowest error.

\subsection{Additional Next-Step Results}
\label{app:additional_ns_results}
Table \ref{tab:comparison_next_step} below gives numerical values for figure \ref{fig:line_next_step}.
These are the same results as in figure \ref{fig:pred_comparison} with results for our combine data set included.
We see that the combined data set has a similar trend to our Navier Stokes data set.

\begin{figure}[h]
    \centering
    \includegraphics[width=0.9\linewidth]{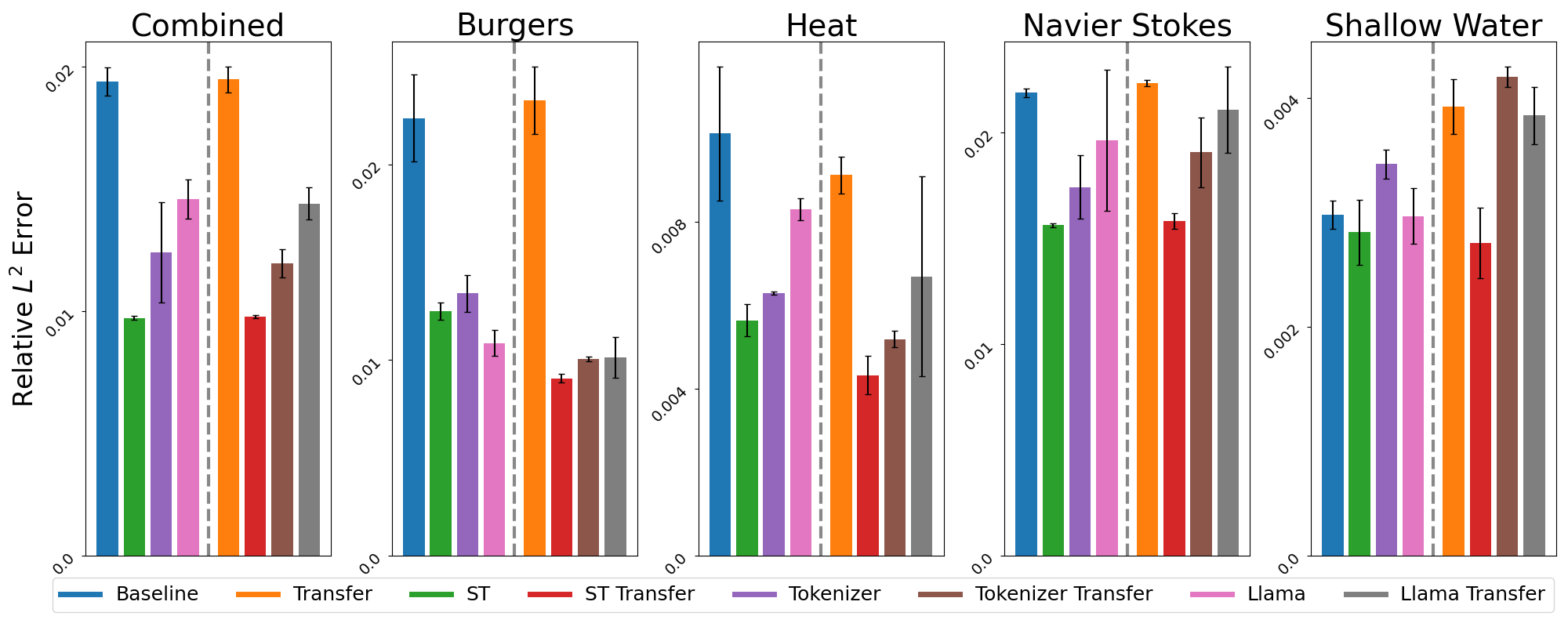}
    \caption{Comparison of next-step prediction relative $L^2$ error for baseline FactFormer, FactFormer + ST, FactFormer + Tokenizer, and Factformer + Llama with and without transfer learning.}
    \label{fig:line_next_step}
\end{figure}
\begin{table}[h]
    \centering
    \caption{Next-Step prediction results ($\times 10^2$}
    \resizebox{\linewidth}{!}{
    \begin{tabular}{c|cccccccc}
        & Baseline & ST & Tokenizer & Llama & Transfer & ST + Transfer & Tokenizer + Transfer & Llama + Transfer  \\
        \hline
Combined & 1.94 $\pm$ 0.06 & \textbf{0.97 $\pm$ 0.01} & 1.24 $\pm$ 0.20 & 1.46 $\pm$ 0.08 & 1.95 $\pm$ 0.05 & 0.98 $\pm$ 0.01 & 1.20 $\pm$ 0.06 & 1.44 $\pm$ 0.07 \\
Burgers & 2.24 $\pm$ 0.22 & 1.25 $\pm$ 0.05 & 1.34 $\pm$ 0.09 & 1.09 $\pm$ 0.07 & 2.33 $\pm$ 0.17 & \textbf{0.91 $\pm$ 0.02} & 1.01 $\pm$ 0.01 & 1.01 $\pm$ 0.10 \\
Heat & 1.01 $\pm$ 0.16 & 0.56 $\pm$ 0.04 & 0.63 $\pm$ 0.00 & 0.83 $\pm$ 0.03 & 0.91 $\pm$ 0.04 & \textbf{0.43 $\pm$ 0.05} & 0.52 $\pm$ 0.02 & 0.67 $\pm$ 0.24 \\
Navier Stokes & 2.19 $\pm$ 0.02 & \textbf{1.56 $\pm$ 0.01} & 1.74 $\pm$ 0.15 & 1.96 $\pm$ 0.33 & 2.23 $\pm$ 0.02 & 1.58 $\pm$ 0.04 & 1.91 $\pm$ 0.16 & 2.11 $\pm$ 0.21 \\
Shallow Water & 0.30 $\pm$ 0.01 & 0.28 $\pm$ 0.03 & 0.34 $\pm$ 0.01 & 0.30 $\pm$ 0.02 & 0.39 $\pm$ 0.02 & \textbf{0.27 $\pm$ 0.03} & 0.42 $\pm$ 0.01 & 0.38 $\pm$ 0.02 \\
    \end{tabular}}
    \label{tab:comparison_next_step}
\end{table}
\clearpage

\subsection{Additional Autoregressive Rollout Results}
Table \ref{tab:comparison_rollout} below gives the numerical values for figure \ref{fig:line_autoregressive_rollout}.
These are the same results as in figure \ref{fig:accumulated_error} with results for our combined data set included.
Again, these results follow the trend of our next-step results closely.

\begin{figure}[h]
    \centering
    \includegraphics[width=0.95\linewidth]{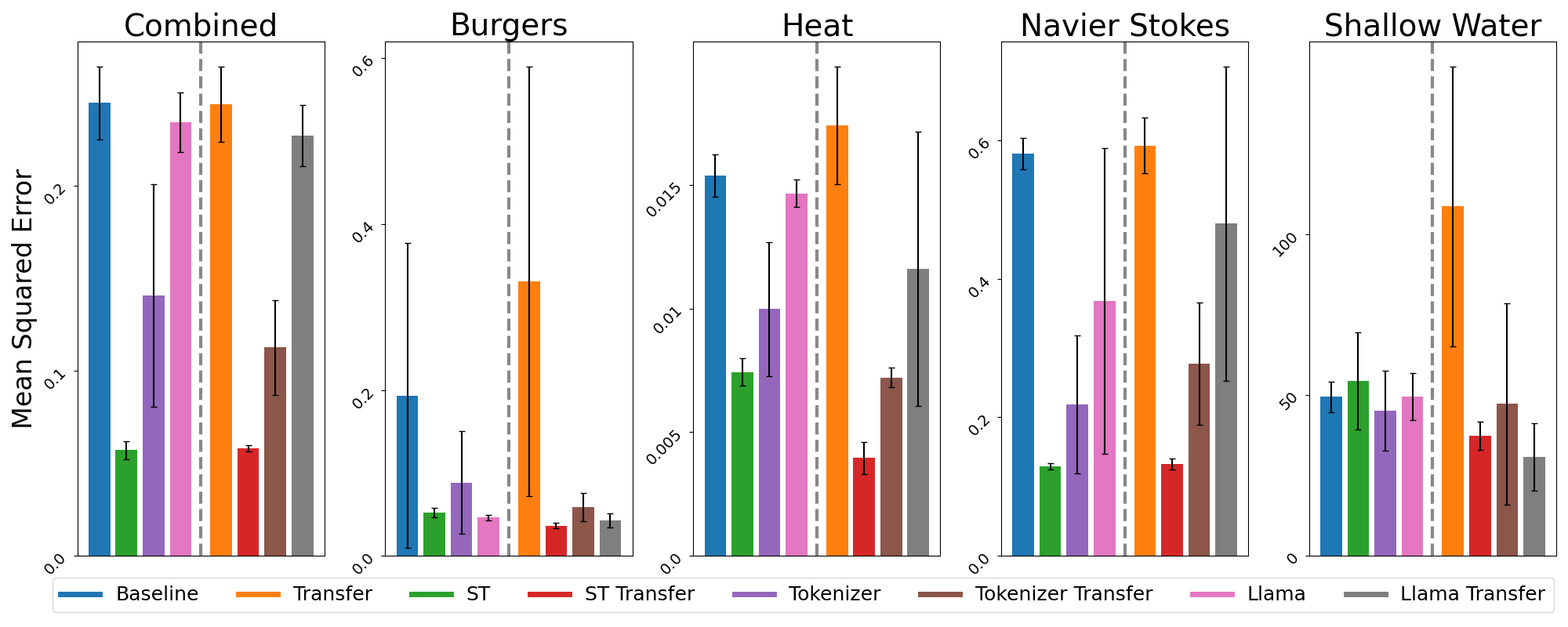}
    \caption{Comparison of accumulated mean squared error for baseline FactFormer, FactFormer + ST, FactFormer + Tokenizer, and baseline + Llama with and without transfer learning.}
    \label{fig:line_autoregressive_rollout}
\end{figure}
\begin{table}[h]
    \centering
    \caption{Accumulated Autoregressive Rollout}
    \resizebox{\linewidth}{!}{
    \begin{tabular}{c|cccccccc}
        & Baseline & ST & Tokenizer & Llama & Transfer & ST + Transfer & Tokenizer + Transfer & Llama + Transfer  \\
        \hline
Combined & 0.24 $\pm$ 0.02 & \textbf{0.06 $\pm$ 0.00} & 0.14 $\pm$ 0.06 & 0.23 $\pm$ 0.02 & 0.24 $\pm$ 0.02 & 0.06 $\pm$ 0.00 & 0.11 $\pm$ 0.03 & 0.23 $\pm$ 0.02 \\
Burgers & 0.19 $\pm$ 0.18 & 0.05 $\pm$ 0.01 & 0.09 $\pm$ 0.06 & 0.05 $\pm$ 0.00 & 0.33 $\pm$ 0.26 & \textbf{0.04 $\pm$ 0.00} & 0.06 $\pm$ 0.02 & 0.04 $\pm$ 0.01 \\
Heat & 0.02 $\pm$ 0.00 & 0.01 $\pm$ 0.00 & 0.01 $\pm$ 0.00 & 0.01 $\pm$ 0.00 & 0.02 $\pm$ 0.00 & \textbf{0.00 $\pm$ 0.00} & 0.01 $\pm$ 0.00 & 0.01 $\pm$ 0.01 \\
Navier Stokes & 0.58 $\pm$ 0.02 & \textbf{0.13 $\pm$ 0.00} & 0.22 $\pm$ 0.10 & 0.37 $\pm$ 0.22 & 0.59 $\pm$ 0.04 & 0.13 $\pm$ 0.01 & 0.28 $\pm$ 0.09 & 0.48 $\pm$ 0.23 \\
Shallow Water & 49.44 $\pm$ 4.73 & 54.28 $\pm$ 15.15 & 45.15 $\pm$ 12.36 & 49.42 $\pm$ 7.35 & 108.57 $\pm$ 43.56 & 37.32 $\pm$ 4.40 & 47.20 $\pm$ 31.23 & \textbf{30.70 $\pm$ 10.47} \\
    \end{tabular}}
    \label{tab:comparison_rollout}
\end{table}
\clearpage

\subsection{Fixed-Future Additional Results}
Table \ref{tab:comparison_fixed_future} below gives the numerical values for figure \ref{fig:line_fixed_future}.
These are the same results as in figure \ref{fig:ff_results} with results for our combined data set included.
We see the combined data set follows very similar trends to the Heat results, where Llama performs best but the other multimodal variants are close in performance.

\begin{figure}[h]
    \centering
    \includegraphics[width=0.9\linewidth]{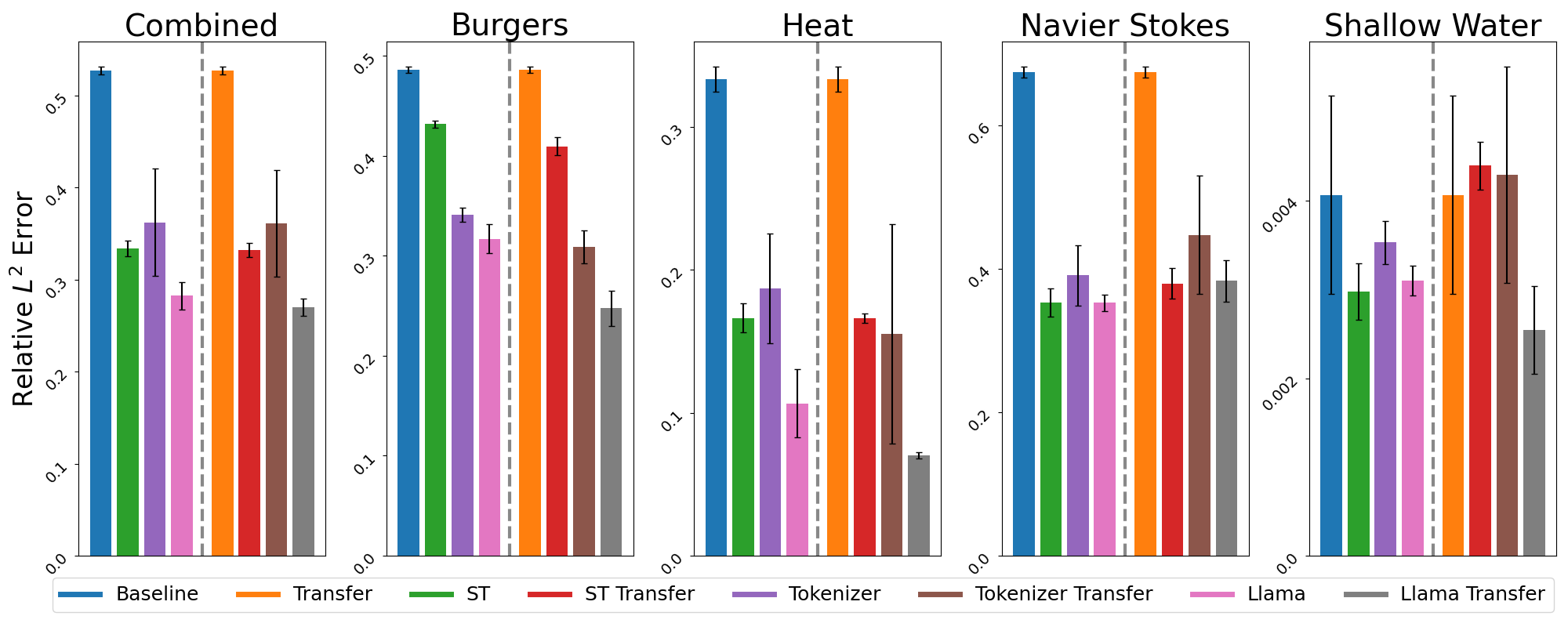}
    \caption{Comparison of fixed-future prediction relative $L^2$ error for baseline FactFormer, FactFormer + ST, FactFormer + Tokenizer, and FactFormer + Llama with and without transfer learning.}
    \label{fig:line_fixed_future}
\end{figure}
\begin{table}[h]
    \centering
    \caption{Fixed-future prediction results}
    \resizebox{\linewidth}{!}{
    \begin{tabular}{c|cccccccc}
        & Baseline & ST & Tokenizer & Llama & Transfer & ST + Transfer & Tokenizer + Transfer & Llama + Transfer  \\
        \hline
Combined & 0.527 $\pm$ 0.004 & 0.334 $\pm$ 0.009 & 0.362 $\pm$ 0.058 & 0.283 $\pm$ 0.015 & 0.527 $\pm$ 0.004 & 0.332 $\pm$ 0.008 & 0.361 $\pm$ 0.058 & \textbf{0.270 $\pm$ 0.009} \\
Burgers & 0.486 $\pm$ 0.003 & 0.431 $\pm$ 0.004 & 0.341 $\pm$ 0.007 & 0.317 $\pm$ 0.015 & 0.486 $\pm$ 0.003 & 0.409 $\pm$ 0.009 & 0.309 $\pm$ 0.016 & \textbf{0.247 $\pm$ 0.018} \\
Heat & 0.334 $\pm$ 0.009 & 0.166 $\pm$ 0.010 & 0.187 $\pm$ 0.038 & 0.107 $\pm$ 0.024 & 0.334 $\pm$ 0.009 & 0.166 $\pm$ 0.003 & 0.155 $\pm$ 0.077 & \textbf{0.070 $\pm$ 0.002} \\
Navier Stokes & 0.674 $\pm$ 0.008 & 0.353 $\pm$ 0.020 & 0.390 $\pm$ 0.042 & \textbf{0.352 $\pm$ 0.011} & 0.674 $\pm$ 0.008 & 0.379 $\pm$ 0.021 & 0.447 $\pm$ 0.083 & 0.383 $\pm$ 0.029 \\
Shallow Water & 0.004 $\pm$ 0.001 & 0.003 $\pm$ 0.000 & 0.004 $\pm$ 0.000 & 0.003 $\pm$ 0.000 & 0.004 $\pm$ 0.001 & 0.004 $\pm$ 0.000 & 0.004 $\pm$ 0.001 & 0.003 $\pm$ 0.000 \\
    \end{tabular}}
    \label{tab:comparison_fixed_future}
\end{table}
\clearpage

\subsection{Ablation Study Additional Results}
\label{app:ablation}
\begin{figure}[h]
    \centering
    \includegraphics[width=0.9\linewidth]{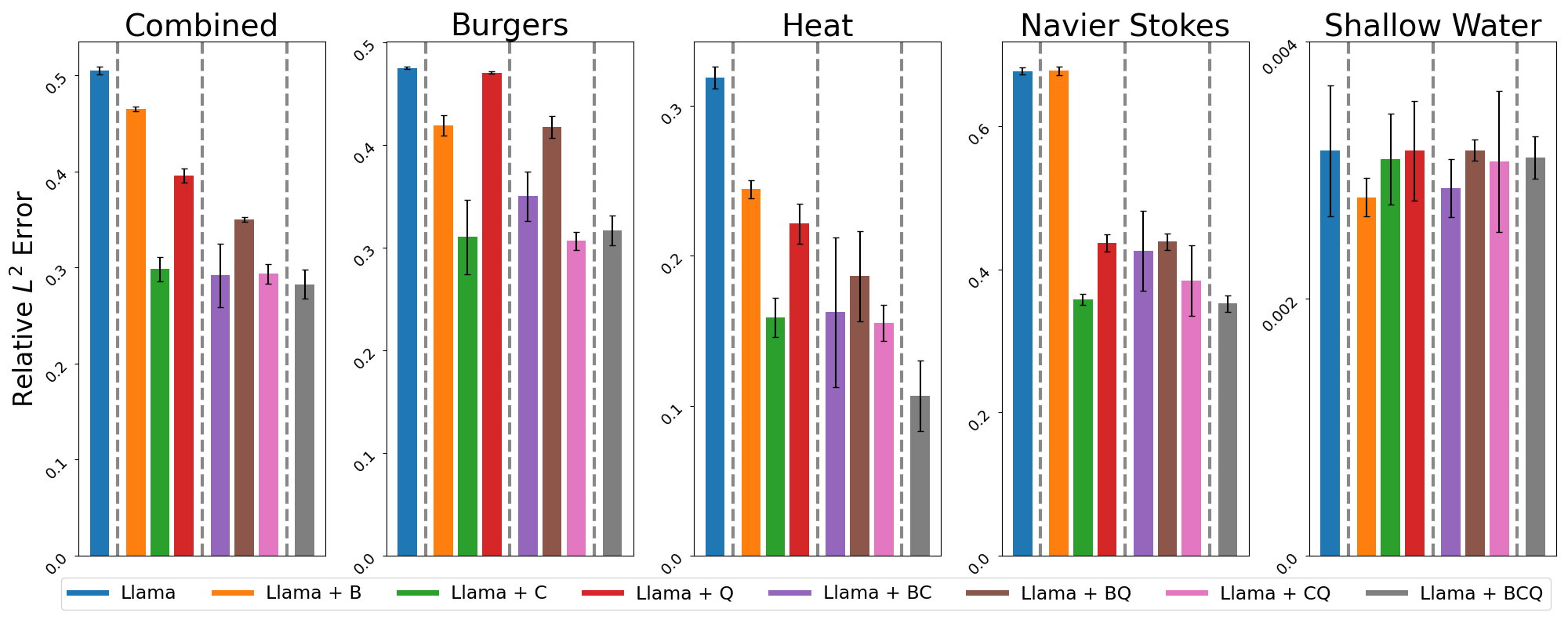}
    \caption{Comparison of fixed-future prediction relative $L^2$ error for FactFormer + Llama with varying levels of sentence description, trained with standard training.}
    \label{fig:line_llama_ablation}
\end{figure}

Table \ref{tab:llama_ablation_next_step} gives the numerical values for figure \ref{fig:line_llama_ablation}.
The ablation study with Llama as the LLM backbone shows very similar to the SentenceTransformer backbone seen in figure \ref{fig:ablation_ff}.
Adding qualitative information improves performance smaller than coefficient information, and only when coefficient information is not also present.

\begin{table}[h]
    \centering
    \caption{Llama Ablation study fixed-future prediction error}
    \resizebox{\linewidth}{!}{
    \begin{tabular}{c|cccccccc}
        & Equation & B & C & Q & BC & BQ & CQ & BCQ \\
        \hline
Combined & 0.505 $\pm$ 0.004 & 0.465 $\pm$ 0.003 & 0.298 $\pm$ 0.013 & 0.395 $\pm$ 0.007 & 0.292 $\pm$ 0.033 & 0.350 $\pm$ 0.003 & 0.294 $\pm$ 0.010 & \textbf{0.283 $\pm$ 0.015} \\
Burgers & 0.475 $\pm$ 0.001 & 0.419 $\pm$ 0.010 & 0.310 $\pm$ 0.036 & 0.470 $\pm$ 0.001 & 0.350 $\pm$ 0.024 & 0.417 $\pm$ 0.011 & \textbf{0.306 $\pm$ 0.009} & 0.317 $\pm$ 0.015 \\
Heat & 0.319 $\pm$ 0.007 & 0.244 $\pm$ 0.006 & 0.159 $\pm$ 0.013 & 0.221 $\pm$ 0.013 & 0.162 $\pm$ 0.050 & 0.186 $\pm$ 0.030 & 0.155 $\pm$ 0.012 & \textbf{0.107 $\pm$ 0.024} \\
Navier Stokes & 0.677 $\pm$ 0.005 & 0.677 $\pm$ 0.006 & 0.358 $\pm$ 0.008 & 0.437 $\pm$ 0.012 & 0.426 $\pm$ 0.056 & 0.439 $\pm$ 0.011 & 0.385 $\pm$ 0.049 & \textbf{0.352 $\pm$ 0.011} \\
Shallow Water & 0.003 $\pm$ 0.001 & 0.003 $\pm$ 0.000 & 0.003 $\pm$ 0.000 & 0.003 $\pm$ 0.000 & 0.003 $\pm$ 0.000 & 0.003 $\pm$ 0.000 & 0.003 $\pm$ 0.001 & 0.003 $\pm$ 0.000 \\
    \end{tabular}}
    \label{tab:llama_ablation_next_step}
\end{table}
\clearpage

\begin{figure}[h]
    \centering
    \includegraphics[width=0.9\linewidth]{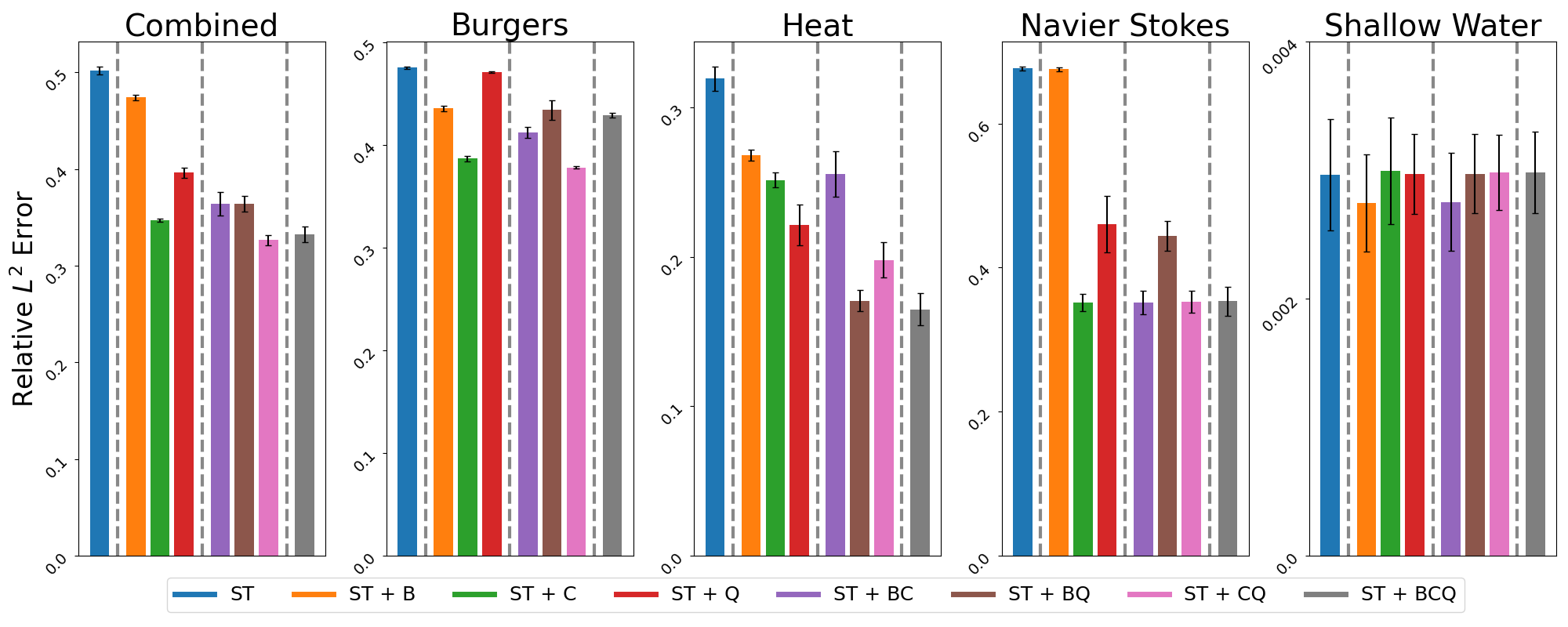}
    \caption{Comparison of fixed-future prediction relative $L^2$ error for FactFormer + ST with varying levels of sentence description, trained with standard training.}
    \label{fig:line_st_ablation}
\end{figure}

Table \ref{tab:st_ablation_next_step} gives the numerical values for figure \ref{fig:line_st_ablation}.
These are the same results as in figure \ref{fig:ablation_ff} with results from our combined data set added.
Again, we see a small improvement in performance when qualitative information is added in tandem with coefficient information in our Burgers and combined data sets, however this is not present in our Navier Stokes data set.
\begin{table}[h]
    \centering
    \caption{ST Ablation study fixed-future prediction error}
    \resizebox{\linewidth}{!}{
    \begin{tabular}{c|cccccccc}
        & Equation & B & C & Q & BC & BQ & CQ & BCQ \\
        \hline
Combined & 0.502 $\pm$ 0.004 & 0.474 $\pm$ 0.003 & 0.347 $\pm$ 0.002 & 0.396 $\pm$ 0.005 & 0.364 $\pm$ 0.012 & 0.364 $\pm$ 0.008 & \textbf{0.326 $\pm$ 0.005} & 0.332 $\pm$ 0.008 \\
Burgers & 0.475 $\pm$ 0.001 & 0.435 $\pm$ 0.003 & 0.387 $\pm$ 0.003 & 0.471 $\pm$ 0.001 & 0.412 $\pm$ 0.005 & 0.434 $\pm$ 0.010 & \textbf{0.378 $\pm$ 0.001} & 0.429 $\pm$ 0.002 \\
Heat & 0.319 $\pm$ 0.008 & 0.268 $\pm$ 0.004 & 0.251 $\pm$ 0.005 & 0.221 $\pm$ 0.014 & 0.256 $\pm$ 0.015 & 0.171 $\pm$ 0.007 & 0.198 $\pm$ 0.012 & \textbf{0.165 $\pm$ 0.011} \\
Navier Stokes & 0.676 $\pm$ 0.003 & 0.675 $\pm$ 0.003 & \textbf{0.352 $\pm$ 0.012} & 0.460 $\pm$ 0.039 & 0.352 $\pm$ 0.016 & 0.444 $\pm$ 0.021 & 0.353 $\pm$ 0.015 & 0.353 $\pm$ 0.020 \\
Shallow Water & 0.003 $\pm$ 0.000 & 0.003 $\pm$ 0.000 & 0.003 $\pm$ 0.000 & 0.003 $\pm$ 0.000 & 0.003 $\pm$ 0.000 & 0.003 $\pm$ 0.000 & 0.003 $\pm$ 0.000 & 0.003 $\pm$ 0.000 \\
    \end{tabular}}
    \label{tab:st_ablation_next_step}
\end{table}

\clearpage

\section{Rollout}
\begin{figure}[h]
    \centering
    \includegraphics[width=0.4\linewidth]{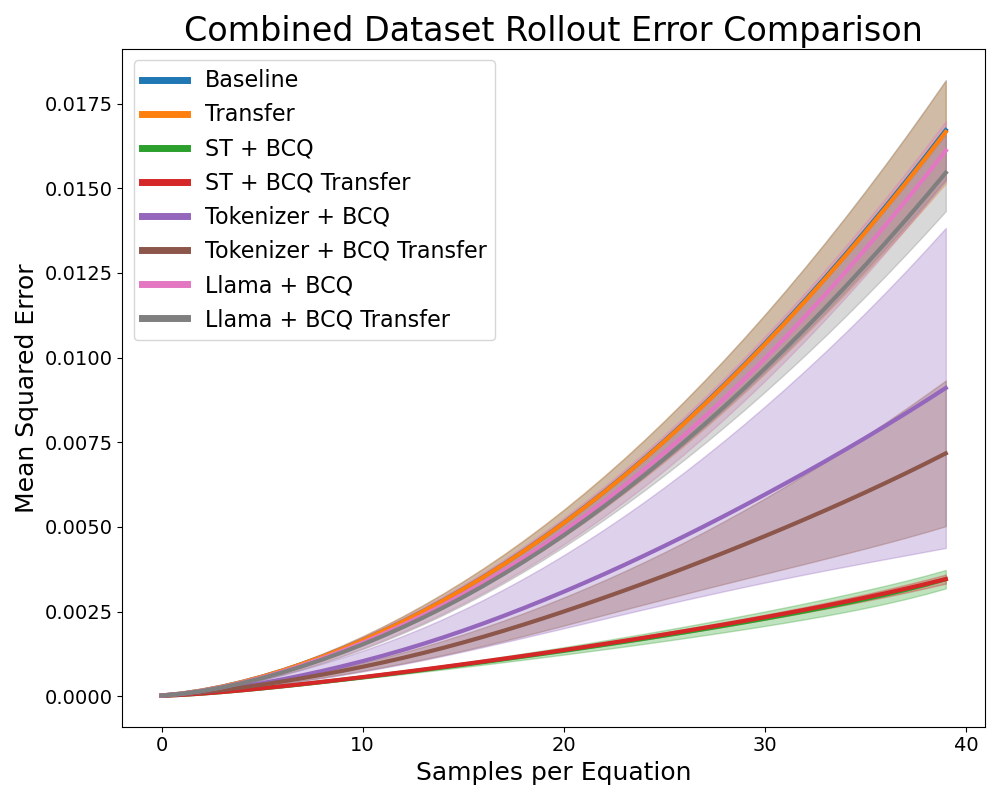}
    \includegraphics[width=0.4\linewidth]{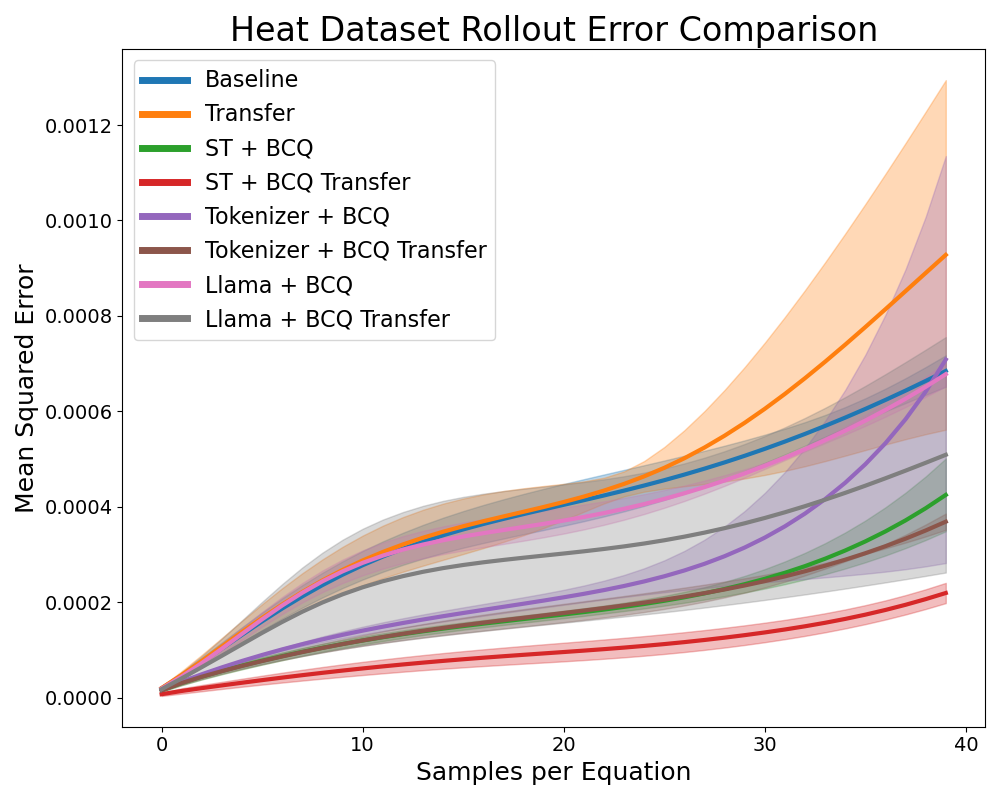} \\
    \includegraphics[width=0.4\linewidth]{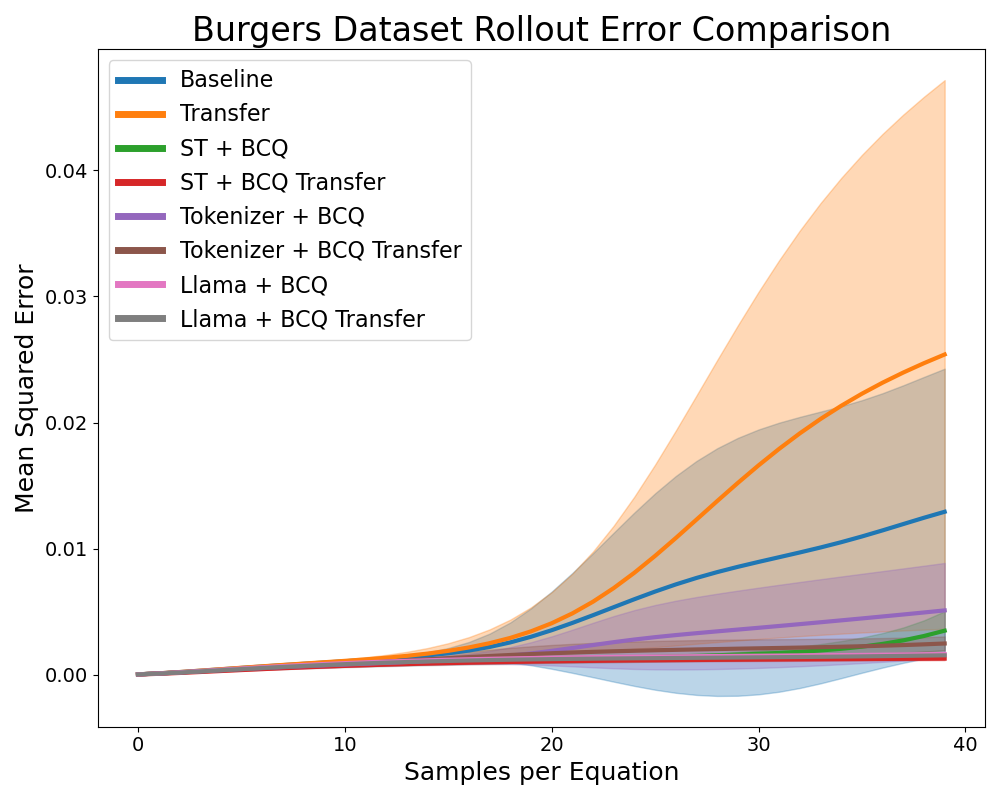}
    \includegraphics[width=0.4\linewidth]{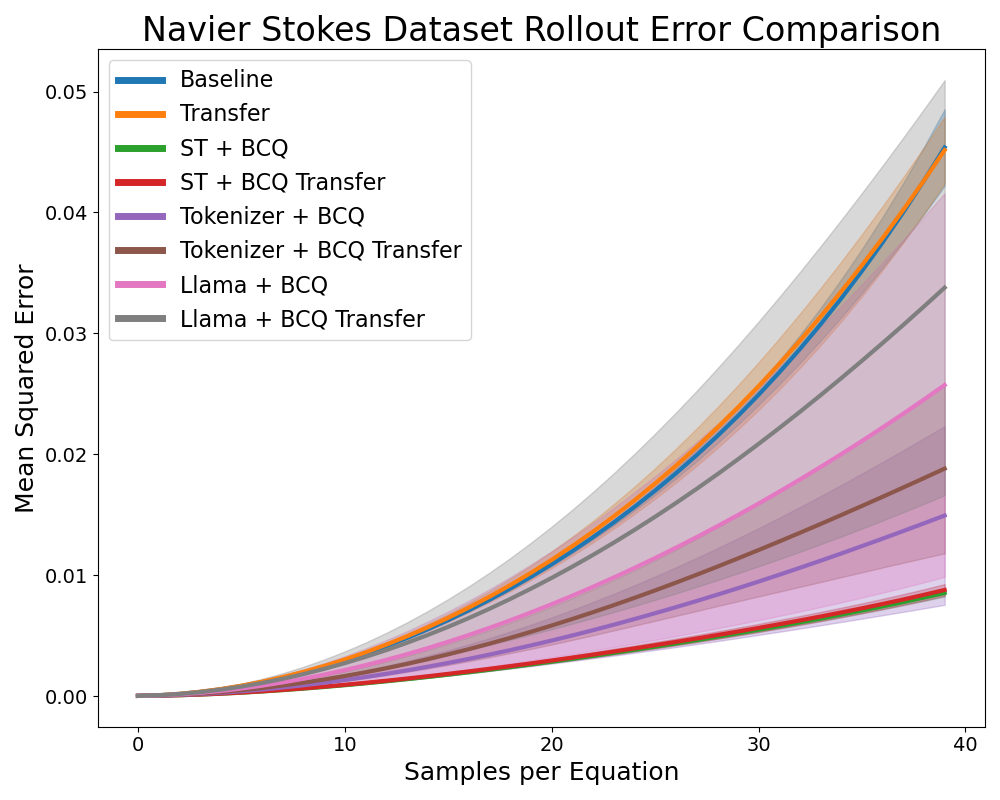} \\
    \includegraphics[width=0.4\linewidth]{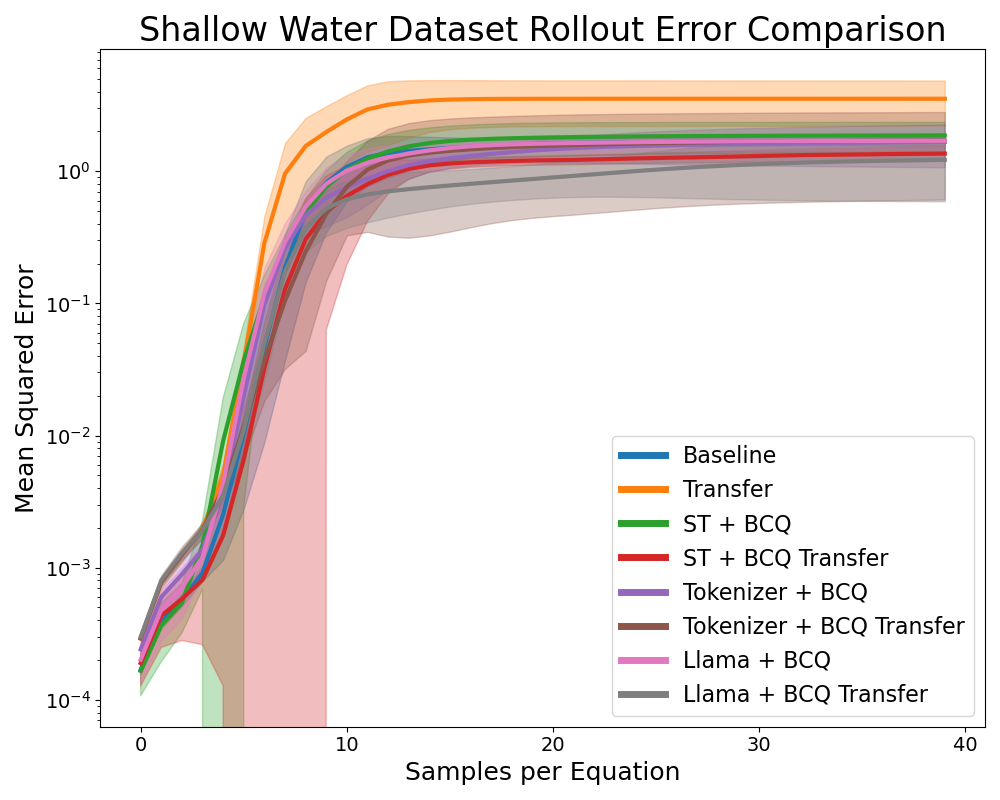}
    \caption{Autoregressive rollout MSE for each dataset.}
    \label{fig:next_step_autoregressive_error}
\end{figure}

Figure \ref{fig:next_step_autoregressive_error} plots the autoregressive rollout error against timestep.
Errors are first averaged over each sample for a given random seed, then the mean and standard deviation across seeds is plotted.

\newpage

\clearpage

%

\newpage

\section{Additional t-SNE Results}
\label{app:tsne}
\begin{figure}[h]
    \centering
    \includegraphics[width=0.8\linewidth]{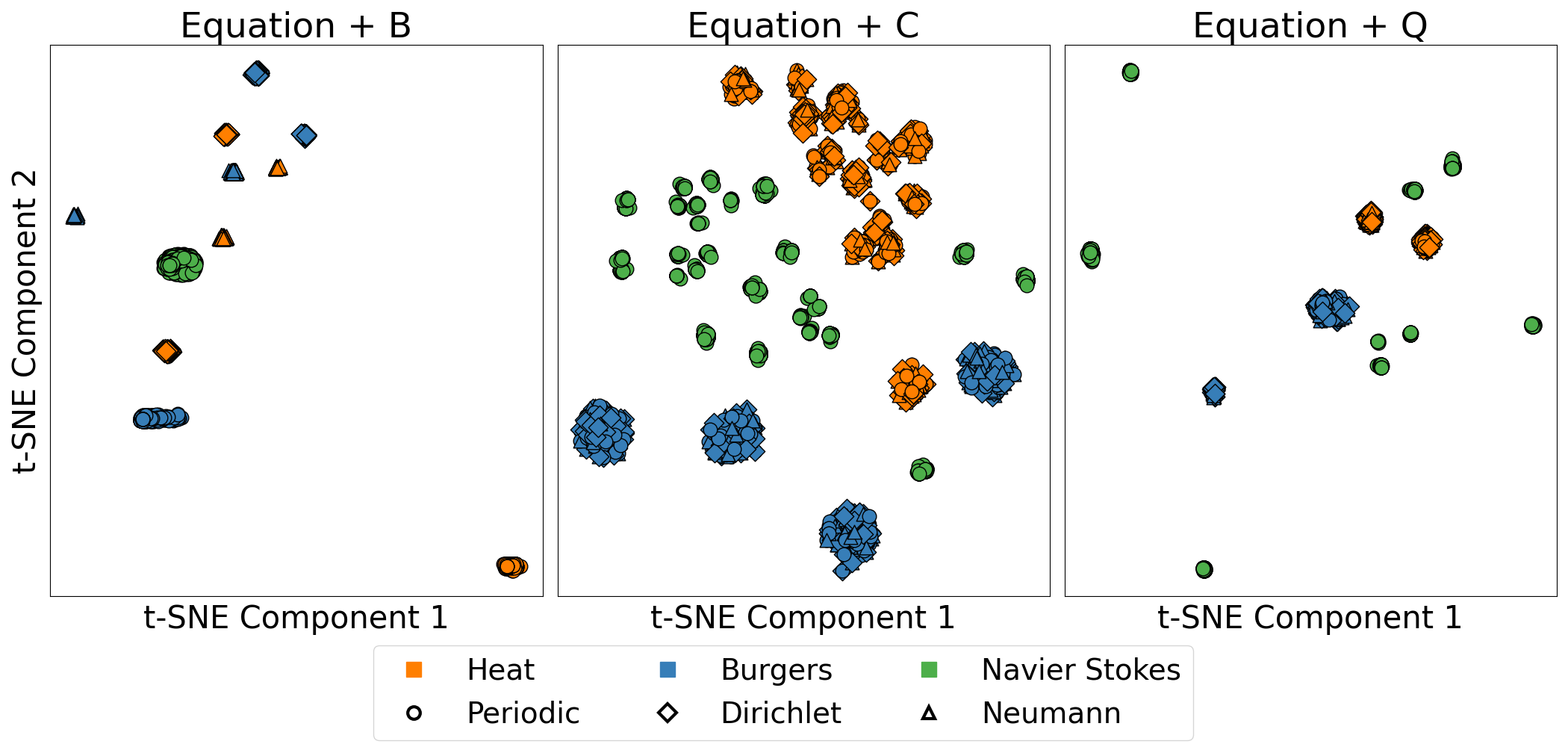}
    \caption{t-SNE embeddings for sentence-level embeddings generated by SentenceTransformer for basic equation description with each of boundary conditions, operator coefficients, and qualitative information separately.}
    \label{fig:llama_single_tsnes}
\end{figure}

\begin{figure}[h]
    \centering
    \includegraphics[width=0.8\linewidth]{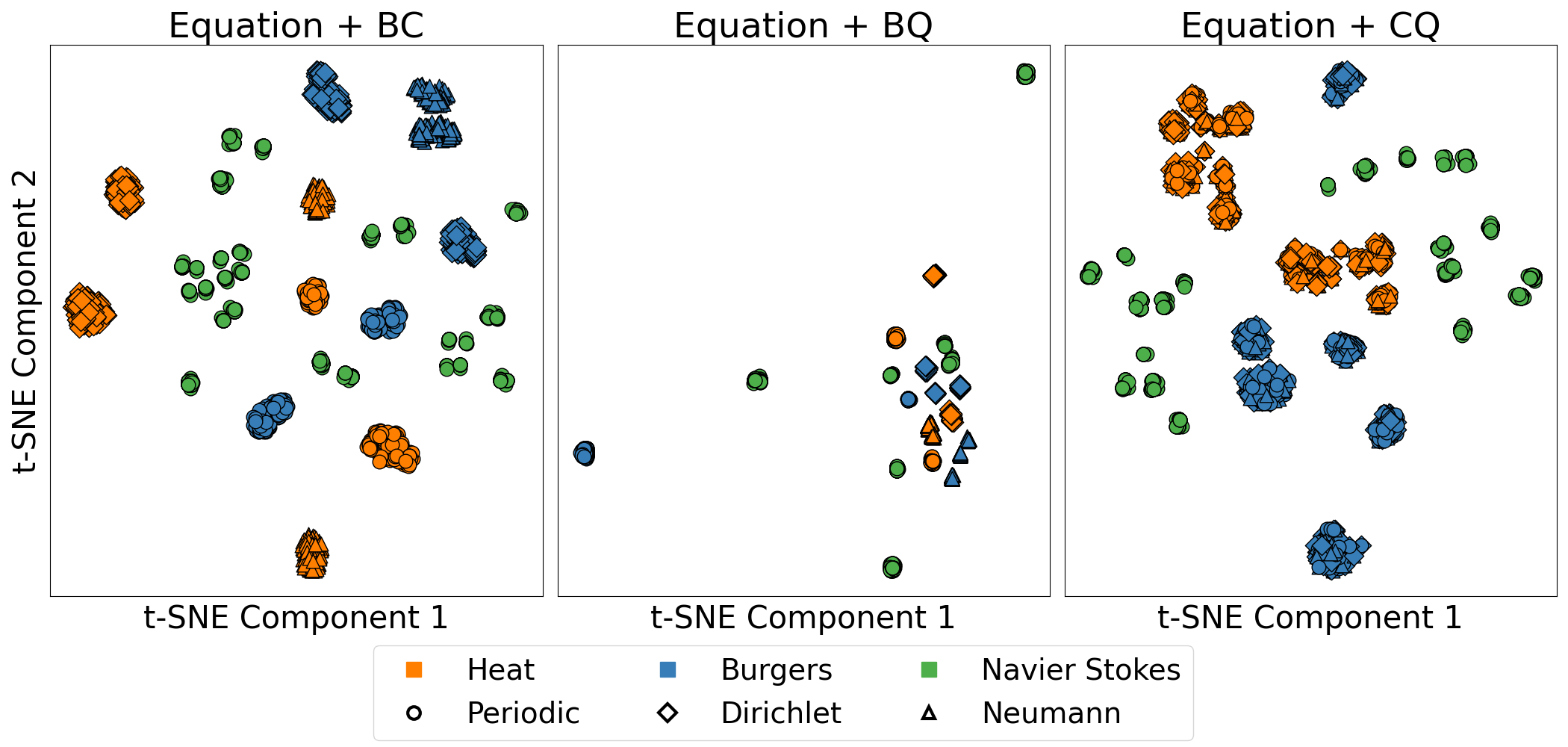}
    \caption{t-SNE embeddings for word-level embeddings averaged over sequence length generated by Llama for each combination of two of boundary conditions, operator coefficients, and qualitative information.}
    \label{fig:llama_double_tsnes}
\end{figure}

\begin{figure}[h]
    \centering
    \includegraphics[width=0.8\linewidth]{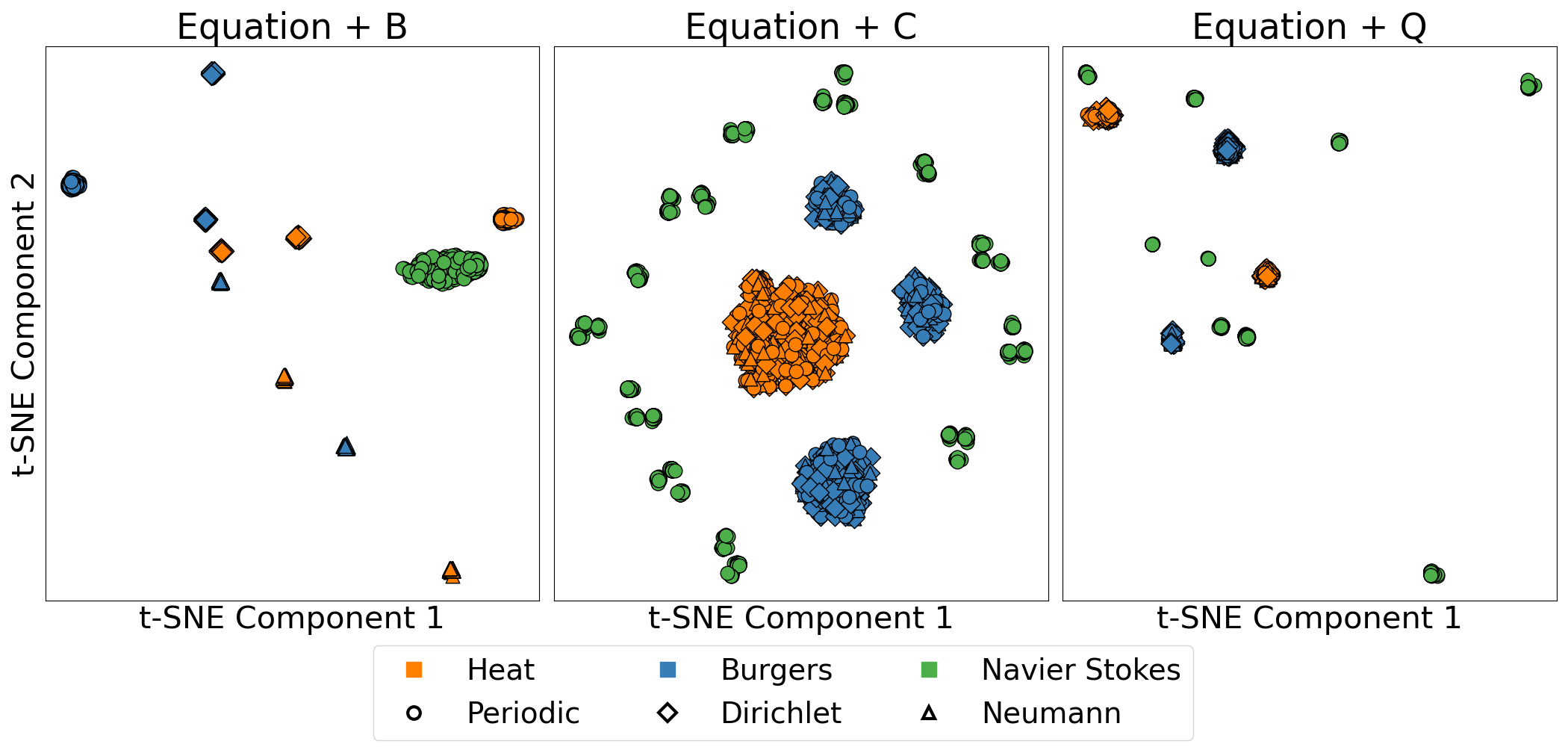}
    \caption{t-SNE embeddings for word-level embeddings averaged over sequence length generated by Llama for each of boundary conditions, operator coefficients, and qualitative information.}
    \label{fig:st_single_tsnes}
\end{figure}

\begin{figure}[h]
    \centering
    \includegraphics[width=0.8\linewidth]{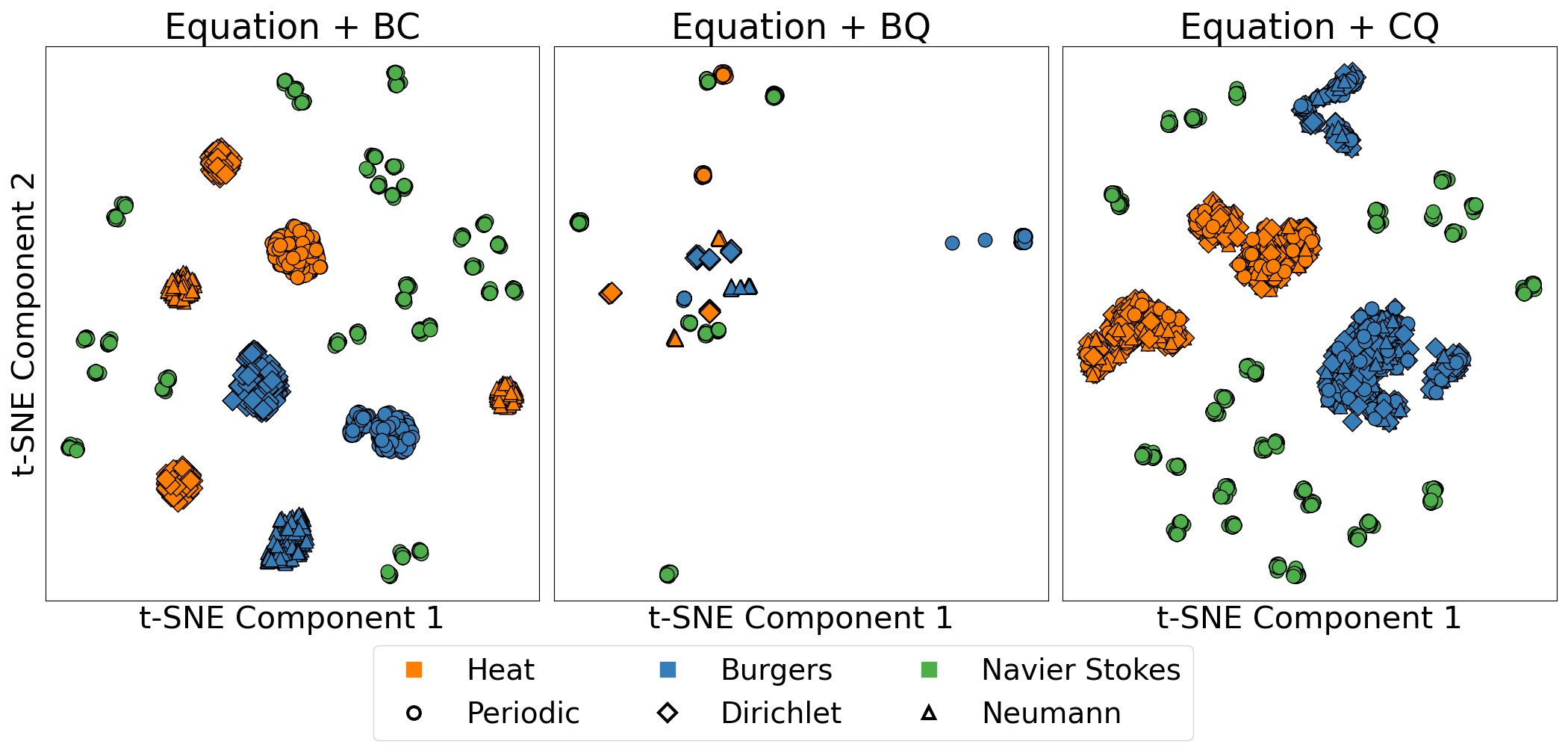}
    \caption{t-SNE embeddings for sentence-level embeddings generated by SentenceTransformer for each combination of two of boundary conditions, operator coefficients, and qualitative information.}
    \label{fig:st_double_tsnes}
\end{figure}

We see as we increase sentence complexity, we get additional structure in our t-SNE embeddings.
Adding coefficient information captures the distribution of coefficients well, adding boundary conditions captures the distinct categories well. 
Using both coefficient information and boundary condition information shows clustering by boundary condition, as well as more clusters than when using qualitative information.
These results are true for both Llama embeddings in figures \ref{fig:llama_single_tsnes} and \ref{fig:llama_double_tsnes}, as well as SentenceTransformer embeddings in figures \ref{fig:st_single_tsnes} and \ref{fig:st_double_tsnes}.

\clearpage

\end{document}